\RequirePackage{fix-cm}
\documentclass[twocolumn]{svjour3}          
\smartqed  
\usepackage{graphicx}
\usepackage{scrtime}
\usepackage[dont-mess-around]{fnpct}
\usepackage{xspace}
\usepackage{natbib}
\usepackage{mathtools}
\usepackage{breqn}
\usepackage{adjustbox}
\usepackage{url}
\usepackage{etoolbox}
\newcommand*{\affaddr}[1]{#1} 
\newcommand*{\affmark}[1][*]{\textsuperscript{#1}}

\newcommand*{\sect}{\S}
\newcommand*{\tab}{Table\@\xspace}
\makeatletter\newcommand*{\etc}{\@ifnextchar{.}{etc}{etc.\@\xspace}}\makeatother
\makeatletter\newcommand*{\eg}{\@ifnextchar{.}{eg}{eg.\@\xspace}}\makeatother
\makeatletter\newcommand*{\ie}{\@ifnextchar{.}{ie}{ie.\@\xspace}}\makeatother
\makeatletter\newcommand*{\wrt}{\@ifnextchar{.}{wrt}{wrt.\@\xspace}}\makeatother
\makeatletter\newcommand*{\vs}{\@ifnextchar{.}{vs}{vs.\@\xspace}}\makeatother
\makeatletter\newcommand*{\fig}{\@ifnextchar{.}{Fig}{Fig.\@\xspace}}\makeatother
\makeatletter\newcommand*{\eq}{\@ifnextchar{.}{Eq}{Eq.\@\xspace}}\makeatother

\newcommand*{\argmin}{\operatornamewithlimits{argmin}}

\journalname{International Journal of Computer Vision}
\begin{document}
\title{Semantic Hierarchical Priors for Intrinsic Image Decomposition}
\titlerunning{Semantic Hierarchical Priors for Intrinsic Image Decomposition}        
\author{Saurabh Saini \affmark[1]  \and  P. J. Narayanan \affmark[1] }
\authorrunning{Saurabh Saini \and P. J. Narayanan} 
\institute{Saurabh Saini \at
	  \email{saurabh.saini@research.iiit.ac.in}   \\
	    \and
	   P. J. Narayanan \at
	  \email{pjn@iiit.ac.in}          \\
	  \affaddr{\affmark[1] International Institute of Information Technology-Hyderabad, Gachibowli, Hyderabad, India - 500032}
}
\date{Received: date / Accepted: date}
\maketitle
\begin{abstract}
Intrinsic Image Decomposition (IID) is a challenging and interesting computer vision problem with various applications in several fields.
We present novel semantic priors and an integrated approach for single image IID that involves analyzing image at three hierarchical context levels.
\textit{Local} context priors capture scene properties at each pixel within a small neighbourhood.
\textit{Mid-level} context priors encode object level semantics.
\textit{Global} context priors establish correspondences at the scene level.
Our semantic priors are designed on both fixed and flexible regions, using selective search method and Convolutional Neural Network features.
Our IID method is an iterative multistage optimization scheme and consists of two complementary formulations: $L_2$ smoothing for shading and $L_1$ sparsity for reflectance.
Experiments and analysis of our method indicate the utility of our semantic priors and structured hierarchical analysis in an IID framework.
We compare our method with other contemporary IID solutions and show results with lesser artifacts.
Finally, we highlight that proper choice and encoding of prior knowledge can produce competitive results
even when compared to end-to-end deep learning IID methods, signifying the importance of such priors. 
We believe that the insights and techniques presented in this paper would be useful in the future IID research.

\keywords{Intrinsic Image Decomposition \and Albedo \and Shading \and Inverse Rendering \and Image Editing}
\end{abstract}
\section{Introduction and Motivation}
\label{sec:intro}
\begin{figure*}[t!]
  \centering
      \begin{tabular}{c c}
      \centering
	  \includegraphics[width=0.48\linewidth,height=2.8cm]{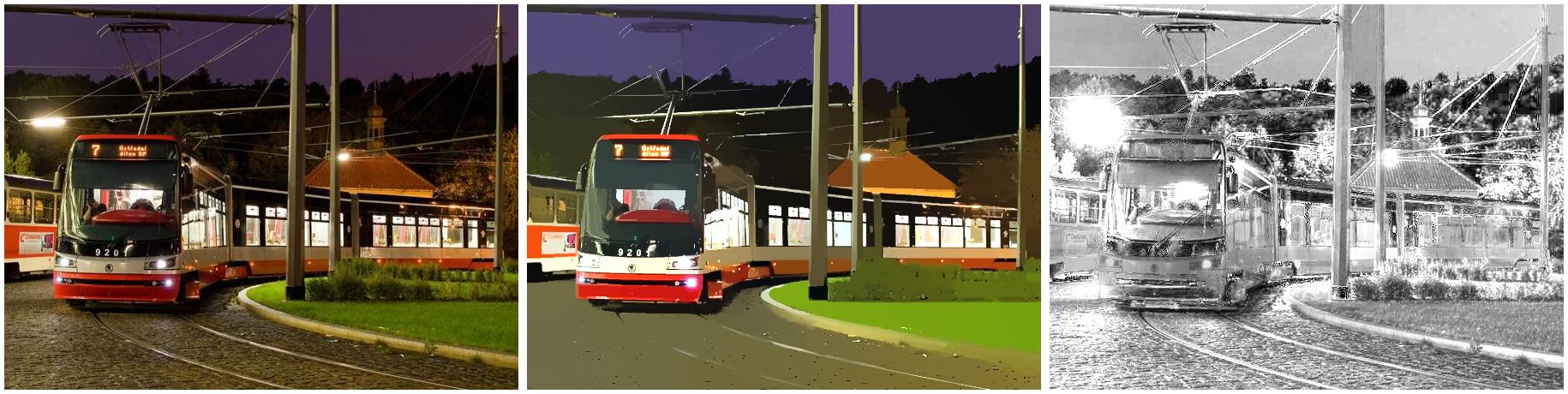} &
	  \includegraphics[width=0.48\linewidth,height=2.8cm]{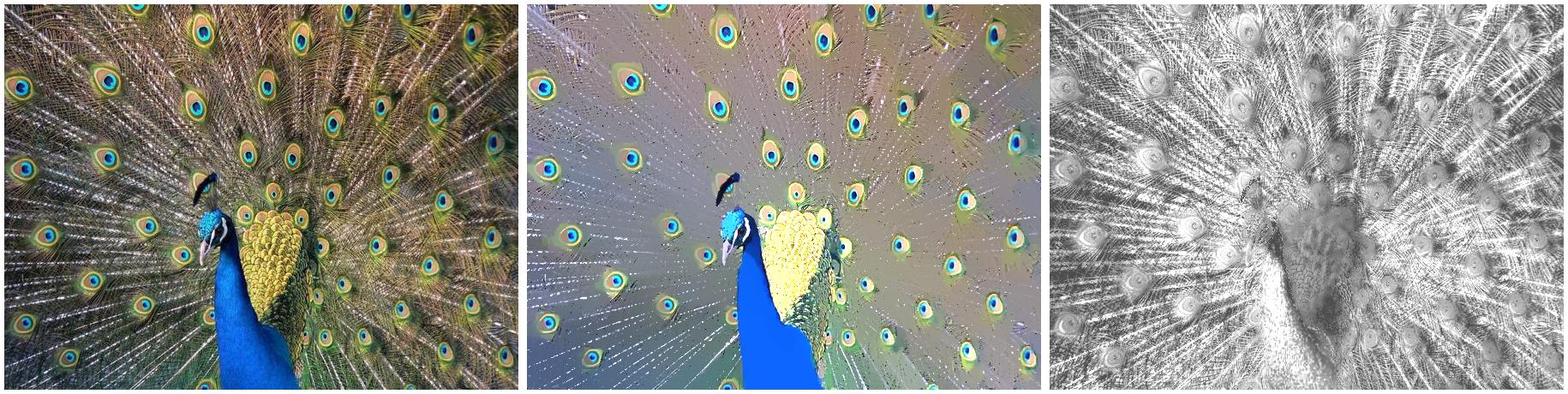} \\
      \end{tabular}
      \caption{\textbf{Intrinsic Image Decomposition (IID):} IID decomposes a given image ($I$) into intrinsic reflectance ($R$) and shading ($S$) components such
      that $I= R\cdot S$ with $R$ containing object colour properties and $S$ capturing scene lighting information. From left to right: $I$, $R$ and $S$ for two images. Notice the colour consistency in the reflectance and separation of lighting and shadows into the shading for the two examples. (All images in this paper best viewed in colour print or as electronic version.)}
      \label{fig:1}
\end{figure*}

Humans are good at visual understanding of several aspects of a scene. 
We can detect and recognize various objects, do semantic associations and guess structural properties in a scene.
We also have the capacity to infer if the visual effects in the image are due object properties or scene lighting.
Intrinsic Image Decomposition (IID) as a research problem is motivated by this observation.
Enabling computers to distinguish light-based and object property-based image effects will improve research in downstream tasks like understanding and image rendering.
This is of interest from both computer vision and computer graphics perspectives.

IID is a classic problem first proposed by \cite{Land71} and studied by both computer vision and graphics research communities.
IID can be categorized under the broad field of research of \textit{inverse rendering} \citep{MarschnerThesis,RamamoorthiInverseRendering} 
which tries to estimate pre-image rendition data of the scene like lighting and albedo, by reversing the light transport models.
In IID we split a given image ($I$) into two underlying components: 
\[
I=R\cdot S, 
\]
where $R$ (reflectance) captures the object dependent properties like colour, textures, \etc
and $S$ (shading) represents direct and indirect lighting in the scene. 
These components could further be reorganized into subparts by using more detailed image formation models
which take into consideration complex optical effects like specular lighting, subsurface scattering, material reflectivity, translucency, volumetric scattering, \etc. Such complex image decompositions might be needed in some specific scenarios but a simple object-lighting dichotomy based definition of IID as stated above, still enables many interesting computer vision applications.
IID is useful in several computer vision and image editing applications like
image colourization \citep{iidColorization}, shadow removal \citep{kwatraShadowRemoval}, 
re-texturing \citep{ramamoorthiRelight}, scene relighting \citep{multiviewOutdoorRelight}, \etc.

From research perspective, IID is an ill-defined and under-constrained problem \citep{balaIID}.
It is ill-defined as in the presence of general real world complex lighting and material reflective properties,
final appearance of an object in an image can not cleanly be separated by using reflectance and shading components only.
Furthermore it is under-constrained as we have to estimate two variables per pixel from a single intensity value from the given image.
Moreover, IID solutions are inherently ambiguous as there can be multiple valid reflectance and shading decompositions differing by a positive scalar multiplicative factor \citep{IIDstarreport}.
All these issues make IID a challenging and interesting research problem.

Previous IID solutions can be categorized under two classes: (\textit{i}) Solutions which assume auxiliary input data from the scene in the form of depth, user annotations, optical flow, multiview, multiple illuminations, photo collections, \etc. and (\textit{ii}) Solutions which work directly on single images and are more dependent on priors and scene assumptions. Our method belongs in the latter category. We utilize weak semantic information from the scene for building novel priors for IID. This is inspired from the observation that scene semantics, even if weak, give us an idea about the underlying scene structure and the object level association between various image pixels. We harness this information to establish constraints between various pixels to tackle the under-constrained IID problem.
We present two simple techniques for weak semantic feature extraction computed on both flexible (segmentation masks) and fixed (overlapping patches) splitting of image regions. We use these features to build priors at three hierarchical contextual scales in our model. 
In summary, three main contributions of this paper are:
\begin{itemize}
 \item We introduce a technique for capturing weak scene semantic information for both fixed and flexible region definitions using CNN and selective search features for IID.
 \item We analyze scene at three context levels: \textit{local context} where optimization weights are based on a small pixel neighbourhood; \textit{mid-level context} which tries to capture object level semantics and \textit{global context} where various regions of the image are linked based on their shared characteristics at the scene level.
  \item We present a new iterative integrated IID framework based on Split-Bregman iterations \citep{splitBregman} using two competing formulations and generate results with fewer artifacts.
\end{itemize}

We perform experiments to analyze the effect of our semantic priors at various context levels and 
illustrate the decompositions generated by our competing formulations over successive iterations.
We evaluate our method on both real and synthetic datasets and show qualitative and quantitative results with respect to the contemporary IID methods.
We believe this is the first IID solution with explicitly encoded semantic priors.
The major takeaway message of this work is that meaningful priors are very useful to solve an ill-posed problem like IID. We present a case for the significance of using domain specific insights like semantics as priors while designing unsupervised solutions for challenging problems like IID.
As datasets, architectures and loss functions are gradually improving, supervised IID methods employing end-to-end deep learning have recently started showing promise. We believe scene semantics based rich and meaningful priors such as ours, will have strong roles on both supervised and unsupervised systems of the future.

\begin{figure*}[t!]
    \centering
    \includegraphics[width=\linewidth,height=8.2cm]{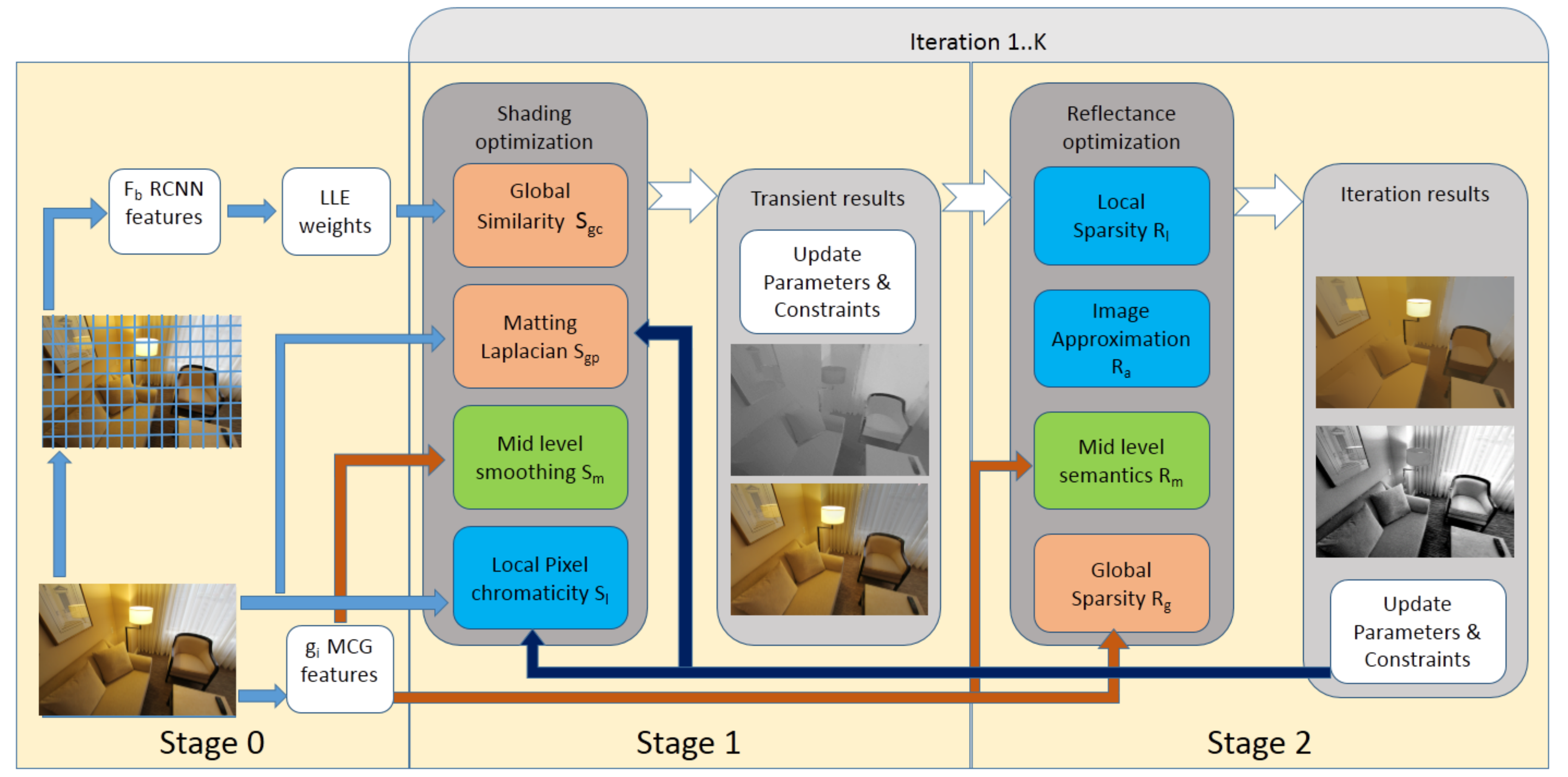}
    \caption{\textbf{Model Outline:} Our method can be understood in three stages. After semantic features extraction (Stage 0), in each iteration our method alternates between the $L_2$ shading (Stage 1) and $L_1$ reflectance optimization (Stage 2) with energy terms computed for both the formulations at three hierarchical context levels: local, mid-level and global.}
    \label{fig:blockDiagram}
\end{figure*}

This paper is based on our previous work \citep{SPIID} with further extensions like additional details, results and visualizations, in all the sections of the paper:
\begin{itemize}
 \item We provide additional details about our motivations in \sect \ref{sec:intro} and more comprehensive report on related IID methods in \sect \ref{sec:relatedWork}.
 \item In \sect \ref{sec:method} we provide additional visualizations and description of our semantic features.
 \item In the analysis section (\sect \ref{sec:Analysis}), we present experiments to justify our design choices and discuss quantitative and qualitative effect of our features and priors.
 \item In the results section (\sect \ref{sec:experiments}), we fine tuned the structure of our model and show additional improvement ($3.27\%$) over our previously reported results. We also present results on two new datasets and evaluate our method using various assessment metrics verifying quality of decomposition.
 \item In \sect \ref{sec:applications} we use the results from our framework to evaluate utility of the method in image editing applications and present two novel image relighting applications changing illumination intensity and colour.
\end{itemize}

\section{Related Work}
\label{sec:relatedWork}

In this section we discuss categories of various IID solutions and other relevant topics in this context pertaining to available datasets, evaluation metrics and supervision type in the learning frameworks.

\paragraph{\textbf{Auxiliary Inputs IID:}}
Several IID methods depend on auxiliary scene data in various forms.
\cite{jeonIID}, \cite{koltunIID} and \cite{barronIID_RGBD} take an RGBD image input and use depth to establish structural correspondences between image pixels.
\cite{bousseauIID} require user annotations in the form of scribbles, marking constant reflectance regions as auxiliary information.
For videos, \cite{videoIID} use optical flow and enforce temporal reflectance consistency constraint between frames.
Similarly \cite{multiviewIID} use multiple views and enforce spatial reflectance consistency by identifying corresponding scene points across images.
This idea is also employed by \cite{weissIID} for reflectance consistency between multiple illumination images.
\cite{fstackIID} base their method on the similar idea and use focal stacks as auxiliary information by substituting it for depth.
\cite{photocollectionIID} and  \cite{iidColorization} use diverse photo collections to establish correspondences between image regions to build constraints.
The common idea behind these methods is to approximate textural and shape similarities using the auxiliary information.
The necessity to acquire additional input data is a major drawback of such methods.

\paragraph{\textbf{Single Image IID:}}
A second category of IID methods work directly on single images.
These methods employ several assumptions and priors, as it is hard to gather sufficient information about geometry, material property and illumination of the scene from a single image. Many such methods work on simple images containing a single object with no background \citep{barronIID_CVPR12, barronIID_PAMI, barronIID_eccv12}. 
Other methods which work on natural scenes utilize priors like \textit{Retinex} \citep{Land71}, reflectance sparsity
\citep{gehlerIID, shenIID}, long \vs short tailed gradient distribution separation \citep{layerSeparation}, spatio-chromatic clustering \citep{clusteringIID}, \etc.
These methods encode interesting insights for the IID problem but are limited when generalizing to `wild' cases with varying lighting and complex textures.
Results vary based on how much significance is given to a prior and the type of optimization framework.
Moreover some of these priors have competing goals. Smoothness prior on shading removes texture details from $S$ as opposed to reflectance sparsity assumption which simplifies
colour details in $R$.
Recent methods try to solve this issue by sequentially employing two separate optimizations for shading and reflectance estimation \citep{balaIID, efrosIID, biIID}.

Based on these insights, our algorithm combines these two types of optimizations in a single integrated algorithm by alternating between two competing formulations:
smoothness for shading and sparsity for reflectance. We use $L_2$ cost terms based optimization for shading and $L_1$ cost terms for reflectance.
We alternate between these two formulations and adapt Split-Bregman iterations for achieving the final decomposition.

\paragraph{\textbf{Auxiliary Outputs IID:}}
Yet another way of categorizing IID solutions is based on the types of outputs generated by these systems.
Some methods instead of assuming the simpler object material \vs scene lighting based dichotomy for IID in the form of reflectance and shading, further divide these components to estimate underlying inverse rendering components like specularity, direct-indirect illumination, specular reflectance, surface normals, \etc. \citet{barronIID_PAMI} estimate shape along with scene intrinsics in their work. \cite{koltunIID} provide direct and indirect lighting as multiplicative shading components. \cite{vibhavIID} present an optimization framework for directly estimating intrinsics, depth, object labels and material attributes from a single RGBD image. 
Similarly \cite{stephenlinIID} and \cite{shelhamerIID} present integrated depth and intrinsics estimation neural network frameworks. 

These methods go one step further while performing IID and enable complex image editing applications, but unlike them we restrict ourselves to the simpler $I=R \cdot S$ formulation and show that the shading and reflectance components from our two stages can be creatively combined to enable several useful image editing applications (\sect \ref{sec:applications}).

\paragraph{\textbf{Datasets:}}
A major challenge associated with IID research is lack of diverse large datasets and proper evaluation metrics \citep{IIDstarreport}.
This arises mainly due to subjective nature of the problem and difficulty in collecting dense annotations. 
MIT intrinsic images dataset introduced by \cite{mitDataset} is limited to a handful of single object images on a black background.
As-Realistic-As-Possible (ARAP) dataset by \citep{IIDstarreport} tries to capture complexity of natural scenes but is also not large enough for supervised training.
Synthetic datasets like MPI Sintel by \cite{mpiSintel}, provide dense annotation but lack sufficient diversity and complexity compared to the natural scenes.
\cite{balaIID} provide a large manually annotated dataset called Intrinsic Images in the Wild (IIW) but have only sparse relative reflectance annotations. 
This limits the utility of such datasets in learning based approaches which aim to work on complete scenes under unrestricted illumination and material property settings. In order to solve this issue several recent deep learning based methods have introduced and designed their systems around such datasets by using time lapse videos \citep{snavelyIID}, synthetically rendered varying illumination scenes \citep{newBiIID}, large scale computer generated images \citep{snavelyIID_ECCV}, \etc. 
This will help advance supervised learning based IID research in the future and we believe that our priors and observations here will be of great assistance while designing such solutions. 

\paragraph{\textbf{Evaluation Metrics:}}
Yet another challenge in IID research is lack of a proper evaluation metric which reflects quantitative, qualitative and applicative performance.
Local Mean Square Error (LMSE) and Structural Similarity Index Metric (SSIM) are used for synthetic scenes and give a sense of mathematical accuracy of the decomposition \citep{koltunIID, jeonIID}. These metrics require dense ground truth annotations and hence difficult to be used on the real images.
\cite{balaIID} suggest a new performance evaluation metric based on their IIW dataset: Weighted Human Disagreement Rate (WHDR).
WHDR gives relative error rate within a given threshold based on the sparse annotations in the IIW dataset. In addition to reflectance evaluation, \cite{sawIID} include manual shading region annotations in the extended IIW dataset to present their Shading Annotations in the Wild (SAW) dataset which can be used to measure shading decomposition accuracy.
As pointed out by \cite{IIDstarreport}, all these metrics do not reflect the applicative ability of IID methods. For this the authors proposed an evaluation strategy based on utilizing IID components in a few automatic image editing applications like logo removal, texture replacement, wrinkles attenuation, \etc. 

Still the lack of a single evaluation metric which could properly evaluate results both perceptually and objectively in various application scenarios, makes comparison between different IID solutions a difficult task. For thorough evaluation, we present results from our method on all these metrics and  also show applicative performance on old and new image editing applications in \sect \ref{sec:experiments} and \sect \ref{sec:applications} respectively.

\paragraph{\textbf{Supervised} \protect\vs \textbf{Unsupervised IID:}}
Some IID methods use supervised learning for IID or to solve related sub-problems like gradient classifiers \citep{tappenIID}, Bayesian graphical models \citep{bayesianNonParamIID} and deep neural networks \citep{efrosIID, narihiraIID, shelhamerIID, stephenlinIID}.
In \cite{efrosIID} and \cite{freemanIID}, authors learn IID priors from Convolutional Neural Network (CNN) using sparse IIW annotations which they propagate to other pixels using a dense Conditional Random Field (CRF) or flood filling the superpixels. 
Other approaches either use the underlying depth information \citep{narihiraIID} 
or use previously proposed RGBD based IID solutions to generate ground truth for supervision \citep{stephenlinIID}. 
Yet another approach is to use dense ground truth from synthetic scenes like from MPI Sintel dataset \citep{mpiSintel} for supervision.
Synthetic datasets like Sintel do not represent true reflectance and shading of natural scenes 
as the dataset was not originally curated with the intention of IID benchmarking \citep{jeonIID}.
Due to domain shift between synthetic \vs real images and limited data, simple CNNs are prone to over-fitting and dataset bias \citep{torralbaDatasetBias, khoslaDatasetBias}.

Recently introduced large synthetic datasets for IID \citep{newBiIID, snavelyIID, snavelyIID_ECCV} have helped in training end-to-end supervised neural network based IID solutions.
These methods are mostly data driven, however explicit encoding of task specific priors can boost their performance.
This inference was also highlighted by \cite{filteringIID} who showed how a simple post processing using guided filtering could improve results of several deep learning IID networks.
Guided or bilateral filtering, which indirectly encodes object level piecewise smoothness prior, has been harnessed to great effect by \cite{filteringIID}, \cite{msrIID} and \cite{newBiIID}.
This suggests that while designing such models, one would be benefited by properly integrating semantic IID priors into their system. 

These issues concerning datasets and performance evaluation, along with it ill-defined nature, make IID a challenging problem to solve using supervised learning frameworks. On the other hand, several older IID methods were unsupervised in nature.
\cite{weissIID}, \cite{gehlerIID}, \cite{vibhavIID} and \cite{bousseauIID} relied on creatively chosen priors and designed their method as optimization schemes.
Such models take advantage of prior knowledge, but either work in restricted settings based upon the assumptions in the models or have scope for performance improvement compared to deep learning based schemes. CNNs have been widely used in computer vision and machine learning literature as black box feature extractor \citep{astoundingCNN, bengioTransferableFeatures, donahueDecaf}. \cite{donahueDecaf} directly use pre-trained CNNs as a feature extractor and prove the generality and cross domain applicability of such features on varied tasks like scene recognition, fine-grained recognition and domain adaptation. Along similar lines, \cite{astoundingCNN} and \cite{bengioTransferableFeatures} also use these features on different tasks and datasets, highlighting their task agnostic characteristics. 

In our model, we absorb the advantages of both the supervised and unsupervised approaches by combining the generality of supervised deep learning methods with prior domain knowledge. We employ an `off-the-shelf' pre-trained deep neural network as a black-box to obtain generic features. 
Additionally we also use an unsupervised method to provide yet another set of semantic features.
We use both these features to introduce new context priors in an unsupervised optimization algorithm by posing it as $L_1$ regularized total variation optimization problem \citep{splitBregman}.
We believe our semantic priors, encode crucial domain insights and would help both supervised and unsupervised future IID solutions.
\section{Method}
\label{sec:method}
Our method is as an iterative algorithm alternating between shading and reflectance formulations (\fig \ref{fig:blockDiagram}).
Optimizing for reflectance sparsity alone leads to loss of textures in reflectance while focusing on shading smoothness leads to 
non-sparse reflectance (see \fig \ref{fig:iterImages}).
We tackle this adversarial nature of the two formulations by estimating IID in two separate stages for shading smoothness and reflectance sparsity.
Such an iterative scheme has earlier also been used by \citet{balaIID} and later adapted by \citet{efrosIID}.
Our framework differs from them as we present a single integrated algorithm without requiring additional steps for building a dense CRF or separate additional optimization frameworks.
We take inspiration from \citet{biIID} who employed \cite{splitBregman}'s Split-Bregman $L_1$-$L_2$ optimization method for image flattening 
and we adapt it to directly estimate IID.
We show that this cleaner integrated approach leads to lesser artifacts in the results while maintaining good quantitative performance.
We discuss these two formulations and the new priors used in our framework below.
\subsection{Semantic Features}
\label{sec:preIteration}

Semantics could provide crucial object and scene information which could help the IID process.
Based on on this intuition, we propose two simple techniques to represent semantic information for IID.
Semantics in images could be either obtained using bounding box annotations or dense segmentation masks.
Both of these problems are separate challenging computer vision research problems in themselves.
Bounding boxes give us weak semantic information whereas dense segmentation masks is a still harder computer vision task and results are often noisier and less accurate compared to the former option. 
In order to avoid solving either of these tasks, we use approximate semantics to build our IID features.
Additionally, this also makes our features class and task agnostic, unlike object detection or segmentation frameworks,
which are limited by the number of classes assumed during training.
This improves the generality of our framework.
We extract two different kinds of features using two complimentary region definitions: fixed and flexible.
We approximate semantic information over these region definitions using two separate techniques as explained below.

\subsubsection{RCNN Features ($f_b$)}
Using the fixed region definition, we divide the input image $I$ into $B$ patches using a fixed grid of a constant size.
In order to extract features from these patches, we pass them through a Region-based Convolutional Neural Network (RCNN) by \cite{rcnn}. 
We pre-train the RCNN on ImageNet dataset \citep{Imagenet} and extract 4096 dimensional features $f_b$ for each patch with $b \in \{1, 2, \ldots B\}$
from the last fully connected layer ($fc7$) of the network. 
We assign this to the center pixel of the patch to obtain a sparse set of regional features for the image.
\citet{convNetsCorrespondence} show that such features, despite having weak label training over the entire scene and large receptive fields,
encode fine correspondences between regions similar to structure encoding features like \textit{SiftFlow} \citep{siftFlow}. 
Hence these features could be used in tasks requiring precise localization like intraclass alignment and keypoints classification.
Furthermore as \textit{SiftFlow} has earlier been used for estimating scene structural information \citep{karschDepth}, 
it provides a good case for applicability of RCNN features for designing IID semantic priors.
Hence we use $f_b$ to approximate shape similarity and estimate correspondences between image patches.
\subsubsection{Selective Search Features ($g_i$)}
\begin{figure*}[t]
\centering
\begin{tabular}{c c c}
  \includegraphics[width=0.3\linewidth,height=3.4cm]{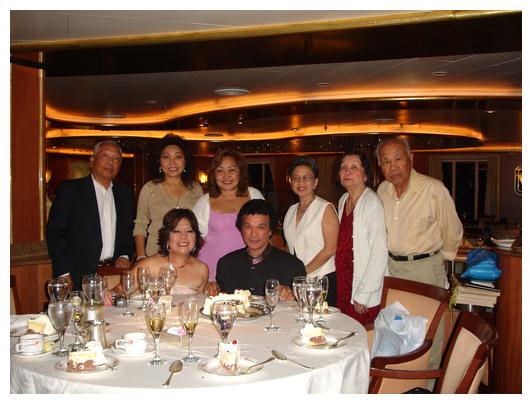} &
  \includegraphics[width=0.3\linewidth,height=3.4cm]{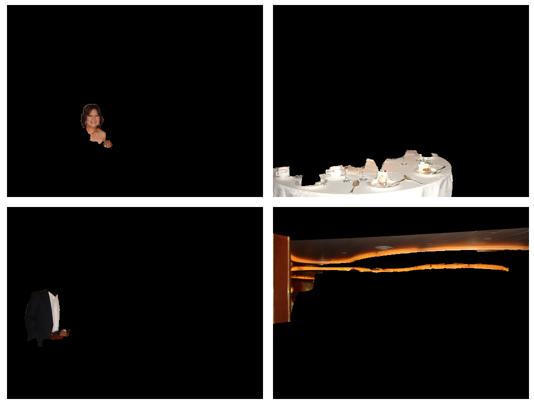} &
  \includegraphics[width=0.3\linewidth,height=3.4cm]{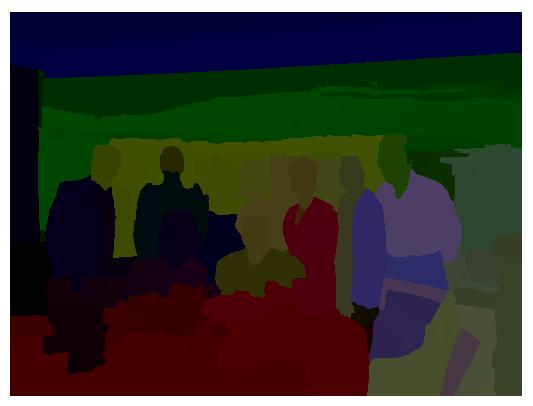} \\
\end{tabular}
\caption{\textbf{Visualizing selective search features.} From left-to-right: Original image; four sample mask images from MCG (binary masks overlaid on the image for visualization) and dimensionality reduced image of selective search features ($g_i$) used for encoding class agnostic semantic information. Notice how regions belonging to an object get grouped together representing approximate semantic information.}
  \label{fig:midContext}
\end{figure*}

Complimentary to fixed patches, we also extract approximate semantic information from flexible region definitions.
Selective search techniques or detection proposals mark interesting image regions which have higher probability of containing an object. 
This improves object detection by avoiding exhaustive sliding window search. Hence selective search results could be used as an indicator of presence of an object (`objectness') in a given region (please see the survey paper by \cite{detectionProposalsSurvey} for further information on selective search techniques).
Selective search is simpler and faster compared to training a Conditional Random Field (CRF) for a finite number of classes for dense pixel associations \citep{biIID, efrosIID}.
Furthermore selective search has off-the-shelf implementations available and does not require separate training.
We use Multiscale Combinatorial Grouping (MCG) by \cite{MCG} for capturing object semantics following the conclusions based on recall and detection quality
from the survey by \citet{detectionProposalsSurvey}.
MCG is a bottom up segmentation method based on fast normalized cuts which are then efficiently assembled into object proposal regions based on an efficient grouping strategy.
MCG generates dense binary region masks and scores for each detection proposal $c \in \{1, 2, \ldots, P\}$ for a total of $P$ proposals.

Our selective search features are formed by concatenating various mask values at a particular pixel, weighted by MCG \citep{MCG} `objectness' score.
We form a concatenated feature vector $g_i$ of proposal masks weighed by proposal score at each pixel and normalize it using $L_2$ norm.
We do dimensionality reduction on these features using PCA for efficient computation during reflectance formulation.
We use dimensionality reduced features in \textit{Stage 2} of the framework unlike \textit{Stage 1} as the mid-level priors are iteratively recomputed only in this stage.
\fig \ref{fig:midContext} shows a few sample masks (overlaid over the image for visualization) and the `PCA-image' (formed by reducing the dimensions to 3) for an example image. Note how in the regions belonging to the same object get clustered together illustrating how our selective search features ($g_i$) encode mid-level semantics. 
\subsection{Shading Formulation}
\label{sec:stage1}
Our shading formulation assumes monochromatic Lambertian illumination and piecewise constant reflectance and is inspired by \cite{jeonIID} which uses depth maps to define pixel neighbourhoods. We generalize their system for a single image by modifying the priors using RCNN and selective search features. The intermediate IID results as shading ($\sigma$) and reflectance ($\rho$), are estimated by minimizing the following energy function:
\begin{equation}
\label{eq:shadingEquation}
\Psi = \lambda_g S_g + \lambda_m S_m + \lambda_l S_l.
\end{equation}
Here $S_g$, $S_m$ and $S_l$ are respectively global, mid-level and local shading priors and $\lambda_g, \lambda_m$ and $\lambda_l$ are the corresponding weights. 

\paragraph{\textbf{Global Context (}$S_{g}$\textbf{):}} Our global shading prior $S_g$ is a combination of a sparse neighbourhood consistency term $S_{c}$ and a weight propagation term $S_{p}$: $S_g = S_c + S_p$.
\cite{jeonIID} show that under the assumption of Lambertian model, shading at a point for a shape can be approximated using a weighted linear combination of surface normals where the weights are computed using Local Linear Embedding (LLE) in the neighbourhood $\mathcal{N}$. As shading is a linear function of surface normals, shading too can be approximated using weighted linear combination of neighbourhood pixel shading values. 
Unlike them we do not have depth information and therefore we approximate structural similarity using the LLE weights of our pre-computed RCNN features $f_b$ as:
\begin{equation}
S_{c}  = \sum_b \big( \sigma_b - \sum_{a \in \mathcal{N}_b } w_{ab}^{c} \sigma_a \big)^2.
\end{equation}
Here $\mathcal{N}_b$ represents the set of 10-nearest neighbours for patch $b$ computed using $f_b$ features and $w^{c}$ are linear combination weights computed using the LLE representation of $b$ over $\mathcal{N}_b$.
These are sparse constraints as we assume the center pixel to be the representative of the entire patch and assign the constraint to it. 
In order to propagate these constraints to the rest of the pixels, we do structure-aware weight propagation using a Laplacian matting matrix \citep{laplacianMatting}.
This approximates shading by an affine function over a base image in a small local window ($\mathcal{N}_{3\times3}$). Our propagation term is defined as:
\begin{equation}
\label{eq:propagationTerm}
 S_{p} = \sum_{i} \sum_{j \in \mathcal{N}_{3\times3}} w_{ij}^{p} \big( \sigma_{i} - \sigma_{j} \big) ^2.
\end{equation}
Here weights $w^{p}$ are computed using the matting Laplacian with reflectance result of the previous iteration as the base image.
For the initial iteration, the base image for the Laplacian is taken as Gaussian smoothened version of $I$.
In their work, \cite{balaIID} propagate global constraints using a dense CRF whereas \citet{efrosIID} devised a Nystr\"om approximation to integrate their proposed CNN reflectance prior for message passing during CRF inference. In comparison, Laplacian matting term has a closed form solution and is easy to compute \citep{jeonIID}.

\paragraph{\textbf{Mid-level Context (}$S_{m}$\textbf{):}} For mid-level prior we use selective search features $g_i$ which encode object semantics.
Similar to the weight propagation term $S_p$, we define this prior as:
\begin{equation}
 S_{m} = \sum_{i} \sum_{j \in \mathcal{N}_{3\times3}} w_{ij}^{m} \big( \sigma_{i} - \sigma_{j} \big) ^2 , \\
\end{equation}
where $w_{ij}^{m} = \exp \big($-$\frac{( 1 - \langle g_i , g_j \rangle)^2}{t_m^2} \big)$ which penalizes dissimilar $g_i$ and $g_j$.
This captures the intuition that in a local neighbourhood if two pixels are predicted to belong to a common object proposal,
then they should have similar shading.
This causes shading smoothness within each detection proposals and preserves texture in the reflectance component.

\paragraph{\textbf{Local Context (}$S_{l}$\textbf{):}} Local context prior is defined following the Retinex model (\ie change in chromaticity implies change in reflectance).
Similar to \cite{jeonIID}, we use this prior in the logarithmic form and substitute $\log{\rho}  = \log{I} - \log{\sigma}$ to obtain:
\[
 S_l = \sum_{i} \sum_{j \in \mathcal{N}_{3\times3}} w_{ij}^{l} \big( (\log{p_i} - \log{\sigma_{i}}) - (\log{p_j} - \log{\sigma_{j}}) \big) ^2 ,\\
\]
where $w_{ij}^{l} = \exp \big($-$\frac{( 1 - \langle \overline{p_i} \cdot \overline{p_j} \rangle)^2}{t_c^2}  \big) \Big [ 1 + \exp \big($-$\frac{p_i^2 + p_j^2}{t_b^2} \big)  \Big] $.
Here $\overline{p_i}$ is pixel chromaticity computed as normalized RGB vector.
The first term in the product awards higher value to similarly coloured pixel pairs. 
The second term gives higher weight to pairs with very low intensity values.
This reduces colour artifacts by suppressing chromatic noise in the dark regions. 
$t_m$, $t_c$ and $t_b$ are fixed deviation parameters for weight estimation.
We solve this quadratic optimization problem ($\sigma^{*} = \argmin_{\sigma}{\Psi}$) using gradient descent and set $\rho^{*} = I-\sigma^{*}$.
\subsection{Reflectance Formulation}
\label{sec:stage2}

Unlike our shading formulation (\sect \ref{sec:stage1}) which enforces smoothness using $L_2$ terms, our reflectance formulation enforces colour sparsity using $L_1$ terms.
The backbone of this stage is inspired from image flattening work by \cite{biIID} which uses Split-Bregman method \citep{splitBregman} for optimization.
For IID, they use flattened image as input and perform a series of steps like self-adaptive clustering, Gaussian mixture modeling, boosted tree classification, CRF labeling and $L_2$ energy minimization.
We show that we can use Split-Bregman iterations for direct IID by using proper context priors and alternating between shading and reflectance formulations.
In addition to being a direct approach, our method is more robust to clustering artifacts (\fig \ref{fig:comparisonResults}).
Our reflectance formulation is given as:
\begin{equation}
 \label{eq:reflectanceEquation}
 \pi = \gamma_g R_g + \gamma_m R_m + \gamma_l R_l + \gamma_a R_a .
\end{equation}
Here $R_g$, $R_m$, $R_l$ and $R_a$ are global, mid-level, local and image approximation terms respectively and $\gamma_g$, $\gamma_m$, $\gamma_l$ and $\gamma_a$ are the associated weights.
We use a similar definition for local and global prior weights ($v^{l}$ and $v^{g}$) and have a fixed deviation parameter ($t$):
\begin{equation}
\label{eq:weights}
 v_{ij} = \exp \Big( - \frac{(\overline{r_i} - \overline{r_j})^2}{2t^2} \Big).
\end{equation}
Here $\overline{r_i}$ is channel normalized CIELab colour value with a suppressed luminance \citep{biIID}.
Note that unlike \citet{biIID}, we re-estimate priors in each iteration which gradually leads to IID directly instead of image flattening.

\paragraph{\textbf{Local Context (}$R_{l}$\textbf{):}} We define local reflectance energy term by enforcing the piecewise local image sparsity like in \cite{biIID}:
\begin{equation}
\label{eq:RA}
 R_l = \sum_i \sum_{j \in \mathcal{N}_{11 \times 11}} v_{ij}^{l} \| R_i - R_j \| _1 = \|\mathbf{Az}\|_1,
\end{equation}
where $R_i$ represents the reflectance to be computed at pixel position $i$.
This term enforces sparsity on reflectance values using local colour information in the form of weights $v_{ij}$ in a ${11 \times 11}$ neighbourhood. 
This term can be rewritten in matrix form by linearizing the colour channels as a single column ($z$) and assembling a block matrix $A$ of associated pixel weights.

\paragraph{\textbf{Mid-level Context (}$R_{m}$\textbf{):}} As $R_l$ enforces sparsity based only on colour similarity in a small local neighbourhood, 
for mid-level context we enforce sparsity at object level using our selective search features ($g_i$).
For ease of computation, we reduce the dimensions of $g$ to get $\mathring{g}$ using PCA and redefine the weights as:
\begin{equation}
 v_{ij}^{m} = \exp \Big( - \frac{(\overline{r_i} - \overline{r_j})^2}{2t^2} \Big) \Big( - \frac{(\mathring{g_i} - \mathring{g_j})^2}{2t^2} \Big). 
\end{equation}
This prior enforces reflectance sparsity at object level which leads to 
colour constancy within an object. This captures object level semantics better compared to the local reflectance sparsity constraints which might lead to over flattening due to ambiguity between edges, textures and noise in an image. The complete mid-level reflectance prior is given as:
\begin{equation}
\label{eq:RB}
 R_m = \sum_i \sum_{j \in \mathcal{N}_{11 \times 11}} v_{ij}^{m} \| R_i - R_j \| _1 = \|\mathbf{Bz}\|_1.
\end{equation}

\paragraph{\textbf{Global Context (}$R_{g}$\textbf{):}} 
The global reflectance prior encodes reflectance similarity at the scene level which is useful in enforcing colour constancy for various instances and occlusion disconnected parts of an object in the scene. We write $R_g$ as:
\begin{equation}
\label{eq:RC}
 R_g = \sum_{i \in Q} \sum_{j \in Q} v_{ij}^{g} \| R_i - R_j \|_1 = \|\mathbf{Cz}\|_1.
\end{equation}
We define $Q$ as the set of representative pixels obtained from each MCG segmentation by ranking all the pixels in a segmentation according to minimum distance from the mean. 

\paragraph{\textbf{Image Approximation (}$R_{a}$\textbf{):}} This term enforces continuity between the two stages by forcing the reflectance estimate from the current stage to be similar to the intermediate reflectance solution from the previous shading formulation stage. We use:
\begin{equation}
\label{eq:RD}
 R_a = \| R_i - \rho\|_2^2  = \| \mathbf{z} - \pmb{\rho^{*}} \|_2^2 = \|\mathbf{D}\|_2^2.
\end{equation}
\subsection{Iterations and Updates}
\label{sec:Optimization}
Using \eq \ref{eq:RA}, \ref{eq:RB}, \ref{eq:RC} and \ref{eq:RD} we can restate \eq \ref{eq:reflectanceEquation} in matrix form as:
\begin{equation}
\label{eq:reflectanceEquationMat}
 \mathbf{\pi} = \| \mathbf{Az} \|_1 + \| \mathbf{Bz} \|_1 + \| \mathbf{Cz} \|_1 + \| \mathbf{D} \|_2^2.
\end{equation}
This is an $L_1$-$L_2$ minimization problem and can be solved by adapting the Split-Bregman iterations \citep{splitBregman}. Split-Bregman method extends the Bregman iterations based Linearized Bregman algorithm for unconstrained optimization problem to a broad range of equality constrained problems. It is especially suited for image processing problems which have large number of constraints. Bregman iterations are used to find the extrema of convex functions and converge very fast compared to Netwon or Gauss-Seidel iterations. \cite{splitBregman} adapt these iterations for $L_1$ regularized problems by decoupling $L_1$ and $L_2$ portions of the energy function and then setting up Bregman iterations using intermediate variables.
We introduce intermediate variables $\mathbf{b}$ and $\mathbf{d}$ for our optimization problem, which reformulates the equation as:
\begin{multline}
\mathbf{z} = \argmin_{\mathbf{z}} \Big( \|\mathbf{D}\|_2^2 + \theta \big( \| \mathbf{d_1 - Az - b_1} \|_2^2 \\
+ \| \mathbf{d_2 - Bz - b_2} \|_2^2 + \| \mathbf{d_3 - Cz - b_3} \|_2^2 \big) \Big)  .
\end{multline}
Here $\theta$ balances the contribution from reflectance sparsity priors \vs prior for shading consistency from previous stage.
We recompute priors after each iteration for the two formulations based on the current values of $\sigma^{*}$ and $\rho^{*}$ and gradually update the contribution of various weighing parameters ($\lambda$, $\gamma$ and $\theta$), increasing the effect of mid-level priors, global priors and the previous solution, while reducing the effect of local priors over the course of iterations. 
It is challenging to decide the convergence of the iterations like in a general Split-Bregman method as there is no IID metric which can give us an estimate of the quality of the iterative decomposition without ground truth.
We cannot directly use reconstruction error as convergence criterion as it does not convey information about the perceptive quality of the decomposition. 
Hence we empirically estimate the total number of iterations ($k=5$) like other model parameters by manually tuning for optimal results over a small subset of images.
\section{Analysis}
\label{sec:Analysis}
\begin{figure*}[t!]
  \centering
    \includegraphics[width=0.8\linewidth,height=3.8cm]{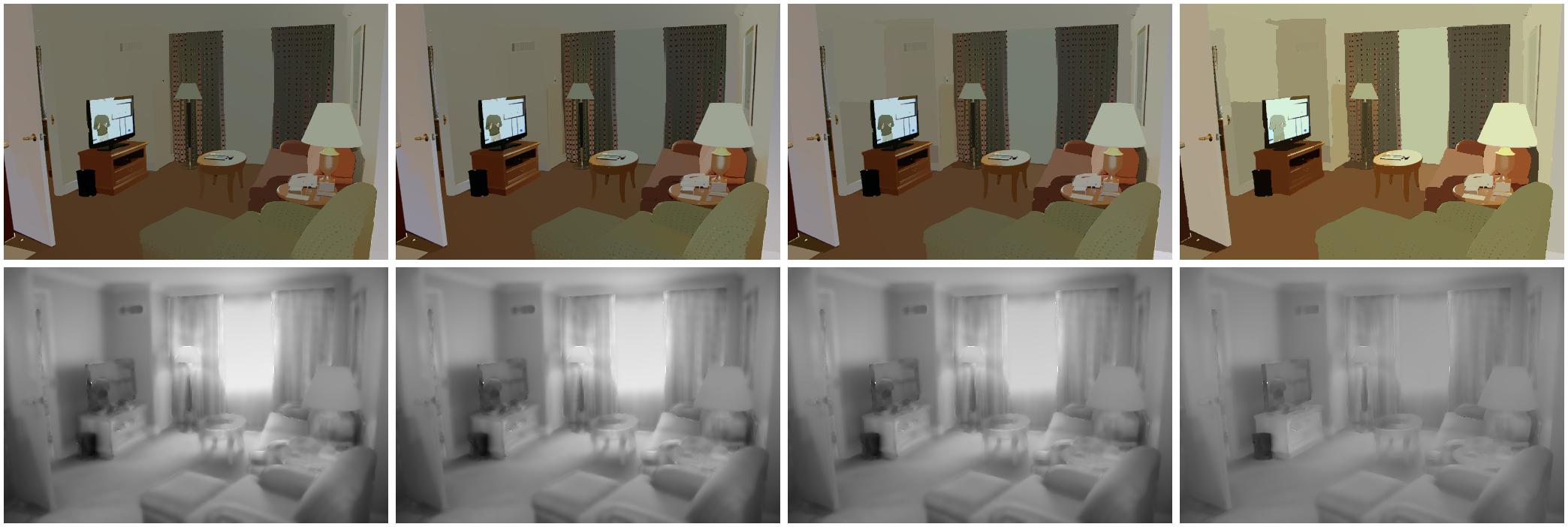}\\
    \includegraphics[width=0.8\linewidth,height=3.8cm]{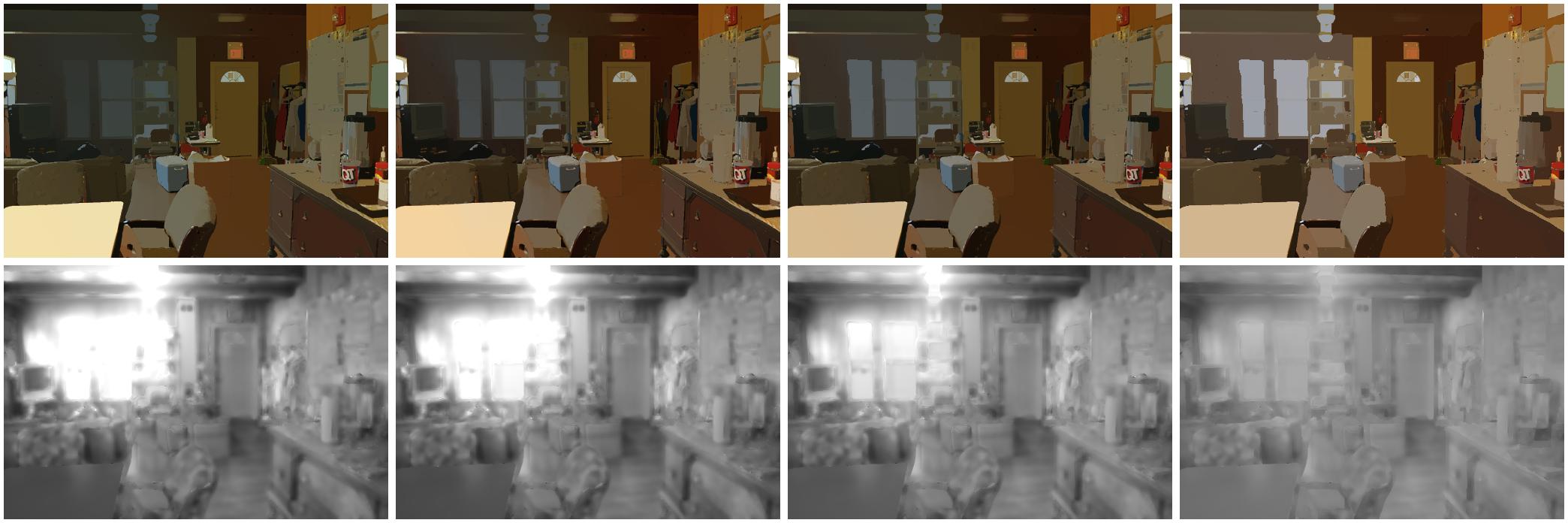}\\
  \caption{\textbf{Framework analysis:} From left to right, shading formulation results ($\sigma^{*}$) and reflectance formulation results ($R^{*}$) for iterations $k = 1, 3, 5, 7$.
  Notice how shading gets `smoother' while reflectance becomes `flatter'.}
  \label{fig:iterImages}
\end{figure*}

\begin{figure}[h]
 \centering
 \includegraphics[width=\linewidth,height=5.0cm]{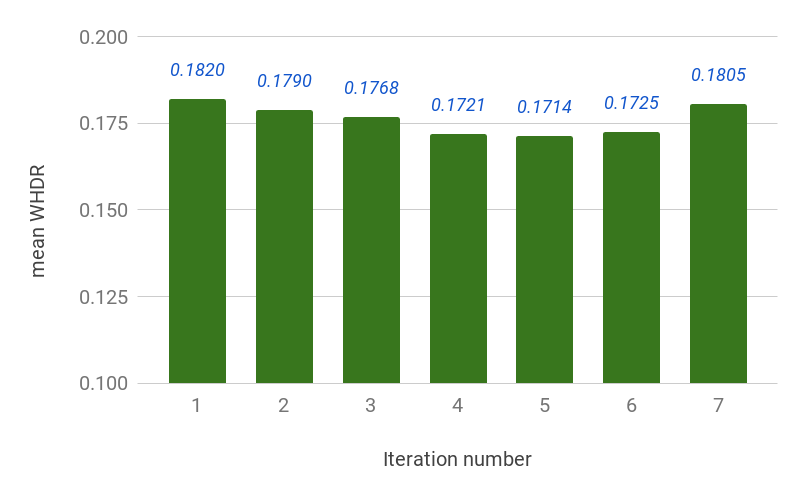}\\
 \caption{\textbf{Iteration analysis:} The graph shows iterative WHDR reduction with minimum at $k=5$.}
 \label{fig:iterGraph}
\end{figure}

\paragraph{\textbf{Feature Analysis:}}
\begin{table}[t!]
\centering
\caption{Experiment results for choosing feature extraction design parameters by varying grid size and overlap region.}
\begin{minipage}{\linewidth}
    \centering
  \begin{adjustbox}{max width=0.98\textwidth}
    \begin{tabular}{c | c || c}
      \multicolumn{3}{c}{\textbf{Fixed grid parameter selection}}  \\
      \hline
      Grid size (pix.) & Stride (pix.) & Mean WHDR \\
      \hline
      \hline
      $30 \times 30$ & $30$ & $18.19$ \\
      $45 \times 45$ & $45$ & $18.09$ \\
      $45 \times 45$ & $30$ & $18.17$ \\
      $60 \times 60$ & $60$ & $18.39$ \\
      $\mathbf{60 \times 60}$ & $\mathbf{30}$ & $\mathbf{17.65}$ \\
      $60 \times 60$ & $15$ & $18.59$ \\
      \hline
    \end{tabular}
  \end{adjustbox}
\end{minipage}
\label{tab:rcnnFeat}
\end{table}  

\begin{table}[t!]
\centering
\caption{Experiment results for various prior computation strategies, with patch based weak semantic features $f_b$ \protect\vs mean appearance based RGB features.}
\begin{minipage}{\linewidth}
    \centering
  \begin{adjustbox}{max width=0.98\textwidth}
    \begin{tabular}{c | c | c || c}
      \multicolumn{4}{c}{\textbf{Semantic features based prior estimation}}  \\
      \hline
      Prior strategy & Feature type & LLE approx. & Mean WHDR \\
      \hline
      \hline
      p1 & RGB   & kNN    & $17.62$ \\
      p2 & RGB   & random & $17.54$ \\
      p3 & $f_b$ & random & $17.42$ \\
      p4 & $f_b$ & kNN    & $\mathbf{17.14}$ \\
      \hline
    \end{tabular}
  \end{adjustbox}
\end{minipage}
\label{tab:rcnnFeatLLE}
\end{table} 

We present results from various experiments which we conducted in order to analyze the effect of varying the design parameters involved during our semantic features extraction stage.
The results from the experiments with varying grid size conducted in order to decide the optimal value for feature computation, are reported in the \tab \ref{tab:rcnnFeat}. 
As we can observe from this table, the size of the grid and the overlap percentage between them, have a significant effect on the overall performance.
Smaller grid don't capture enough contextual information whereas large grids are too ambiguous. 
Similarly too much overlap leads to most of the nearest neighbours getting picked from the same region, reducing the patch diversity and ability of the system to establish global constraints. Following our empirical observations, we used the $60 \times 60$ grid size with a sliding window stride of $30$ for feature extraction.

We also experimented with four different strategies for our global prior term computation (p1,p2,p3 and p4).
We estimate the effect of using our weak semantic features $f_b$ \vs normal RGB appearance based cues.
Additionally, we also analyze the effect of establishing constraints based on LLE approximations computed using k-nearest neighbours (kNN) or randomly chosen patches.
The results of these experiments are shown in \tab \ref{tab:rcnnFeatLLE}.
As can be observed from the \tab \ref{tab:rcnnFeatLLE} using mean RGB value based features alone in place of weak semantic features $f_b$ gives higher error score. Also as RGB values only capture appearance cues and might not indicate correct structural similarity, even randomly choosing patch neighbours (p2) performs better than kNN based linear approximation strategy (p1). RCNN based weak semantic features are able to capture the structural similarity much better than only mean RGB values with kNN strategy improving performance (p4) over random chosen neighbours strategy (p3).

\paragraph{\textbf{Framework Analysis:}}
In \fig \ref{fig:iterImages} we show qualitative performance of our method for a sample image over successive iterations.
Notice how as per the intended design of our framework, reflectance component from our second formulation gradually gets more `flattened' while shading from the first formulation becomes smoother.
Split-Bregman method uses reconstruction error as the stopping criterion \citep{splitBregman,biIID} but in our case it cannot be directly used to quantify IID performance because of unavailability of ground truth at runtime. Hence we empirically estimate the value of $k$. Considering various scene and lighting settings we observed that overall our algorithm achieves peak perceptual and quantitative performance for $k=5$ which can be seen in the WHDR \vs iterations graph in \fig \ref{fig:iterGraph}. Better performance could be obtained if IID quality could be approximated for each image separately without ground truth information. But devising such a metric is non-trivial and beyond the scope of this paper.
From our experiments we observed that manually selecting optimum $k$ for each image separately can reduce the error.


\begin{table}[t!]
\centering
\caption{Ablation study using various variants of our proposed method showing the utility of our hierarchical priors.}
\begin{minipage}{\linewidth}
  \begin{adjustbox}{max width=0.98\textwidth}
    \begin{tabular}{c | c | c || c}
      \multicolumn{4}{c}{\textbf{Ablation Analysis}}  \\
      \hline
      Variant & Shading priors & Reflectance priors & Mean WHDR \\
      \hline
      \hline
      v1 & $S_l		$ & $R_l + R_m + R_g$ & $24.34$ \\
      v2 & $S_l + S_g	$ & $R_l + R_m + R_g$ & $18.70$ \\
      v3 & $S_l + S_m	$ & $R_l + R_m + R_g$ & $17.16$ \\
      v4 & $S_l + S_g + S_m$ & $R_l	    $ & $22.12$ \\
      v5 & $S_l + S_g + S_m$ & $R_l + R_g   $ & $22.09$ \\
      v6 & $S_l + S_g + S_m$ & $R_l + R_m   $ & $16.88$ \\
      \hline
      $\mathbf{v7}$ & $\mathbf{S_l+S_g+S_m}$ & $\mathbf{R_l+R_m+R_g}$ & $\mathbf{16.86}$ \\
      \hline
    \end{tabular}
  \end{adjustbox}
\end{minipage}
\label{tab:ablationAnalysis}
\end{table}   

\paragraph{\textbf{Ablation Study:}} In order to highlight the significance of various context priors, we conducted an ablation study (\tab \ref{tab:ablationAnalysis}) using different variants of our framework formed by combining different prior terms on a set of randomly chosen $500$ IIW images. 
Variant v1 is essentially iterative Retinex model based smoothing followed by image flattening. 
Similarly v4 is only local $L_1$ flattening performed on top of $L_2$ shading formulation.
Addition of other context priors on top of these basic variants successively improves the performance proving the significance of these priors.
In v2 and v5, we introduce the global context priors, leading to improvement in performance over v1 and v4 respectively.
The large error drop from v1 to v2 is due to our global semantic priors based on RCNN features ($f_b$) computed on a fixed grid.
In v3 and v6 we introduce mid-level context priors using selective search features ($g_i$) computed using flexible regions,
which further leads to significant error reduction.
This shows the utility of our semantic priors at various context levels.
Overall the combination of all these priors gives the best IID results which can be observed from comparisons from v2 and v6 \vs v7 which gives the best qualitative and quantitative performance.

\begin{figure*}[t!]
\centering
\includegraphics[width=1\linewidth,height=2.6cm]{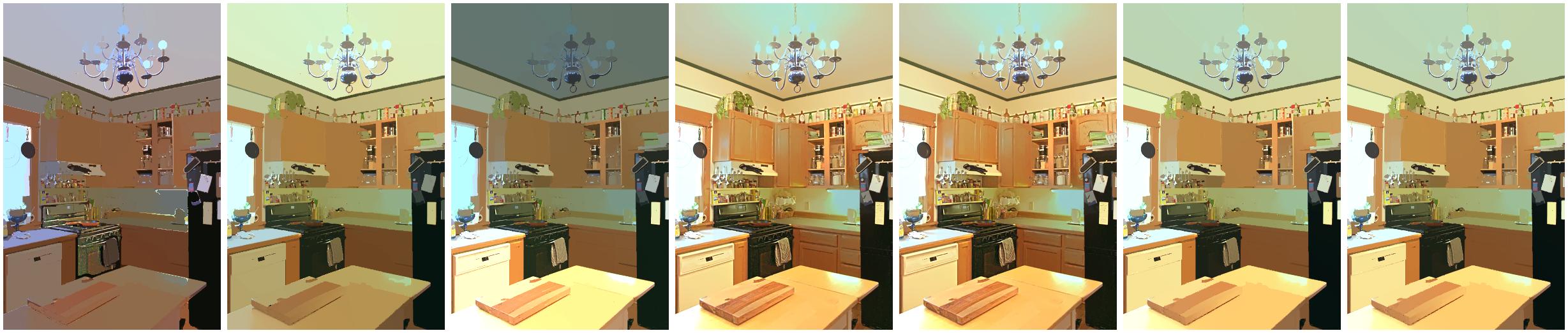} \\
\includegraphics[width=1\linewidth,height=2.6cm]{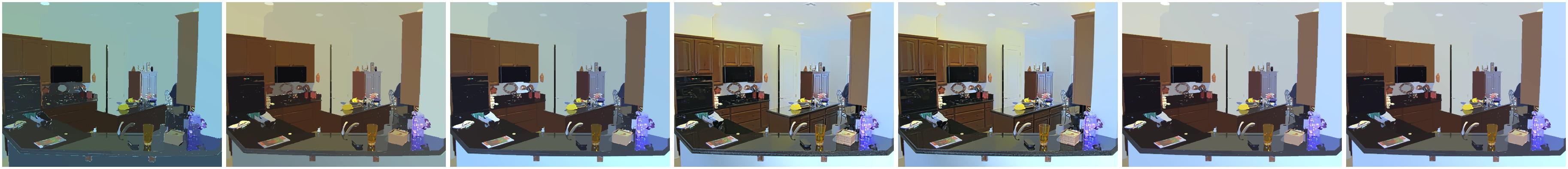} \\
\caption{\textbf{Ablation study:} In each scene from left-to-right: Results using variant v1, v2, v3, v4, v5, v6 and v7.
v3 and v6 give good results as they contain all except one prior. 
v4 and v5 lack mid-level reflectance sparsity term and is unable to remove the highlights from the scene (in $4^{th}$ and $5^{th}$ columns light gradients and shadows are not properly removed). Finally v7 gives overall the best qualitative and quantitative results.}
\label{fig:ablationResults}
\end{figure*}

The qualitative results obtained using these variants are shown in \fig \ref{fig:ablationResults}. 
Note, v1 has very little structural information as most of the shading priors are missing and hence derives results mainly based on colour information. 
This causes incorrect IID reflectance as shown in column 1. 
v2 brings scene level structural information in the form of $S_g$ but in a few cases is unstable as no mid-level semantic information is present.
v3 gives significantly better results compared to previous two as it has nearly all the priors but for a few cases might lead to incorrect global reflectance tone due to lack of global shading information.
v4 and v5 give good reflectance results but do not handle shadows and lights well and contain some artifacts.
These are better handled by v6 due to our semantic prior $R_m$. Finally v7 though looks similar to v6 but also gives overall best quantitative performance.

We reuse the values of most of the parameters in Split-Bregman iterations as provided by \cite{biIID} and empirically estimate the remaining parameters over a small subset of images. All analysis and results in our paper are generated using these fixed set of parameter values:
$\lambda_g = \lambda_m = 0.02$, $\lambda_l = \gamma_g = 2$, $\gamma_m = \gamma_l = 20$, $\theta = 40$, $\tau = 1.2$, $t_c = 0.0001$, $t_m = t_b = t = 0.05$, $k = 5$

\section{Results}
\label{sec:experiments}

\begin{figure*}[t!]
\centering
    \begin{minipage}{0.49\linewidth}
  \includegraphics[width=1\linewidth,height=2.4cm]{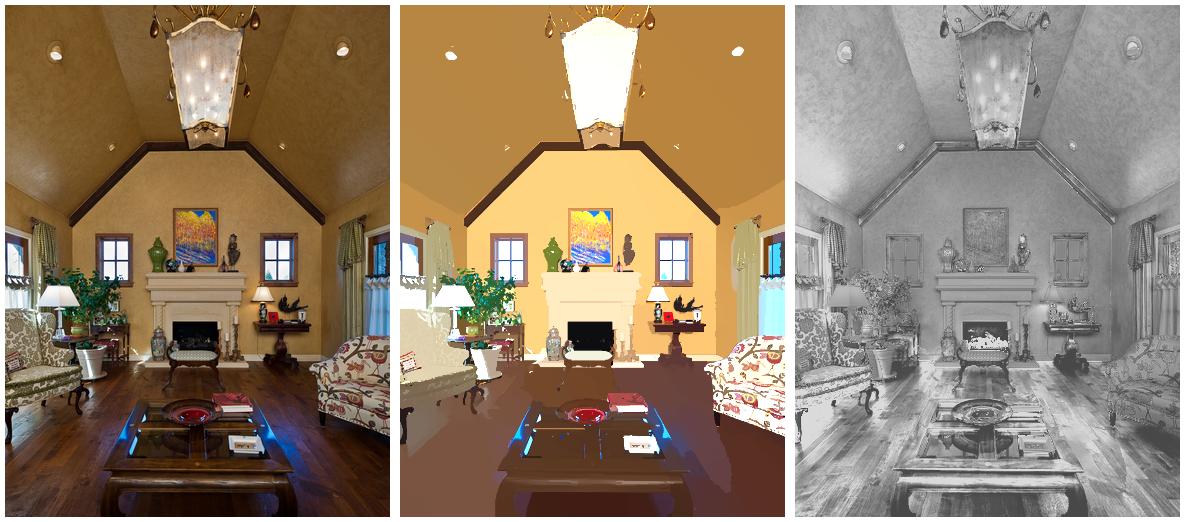} \\
  \includegraphics[width=\linewidth,height=2.4cm]{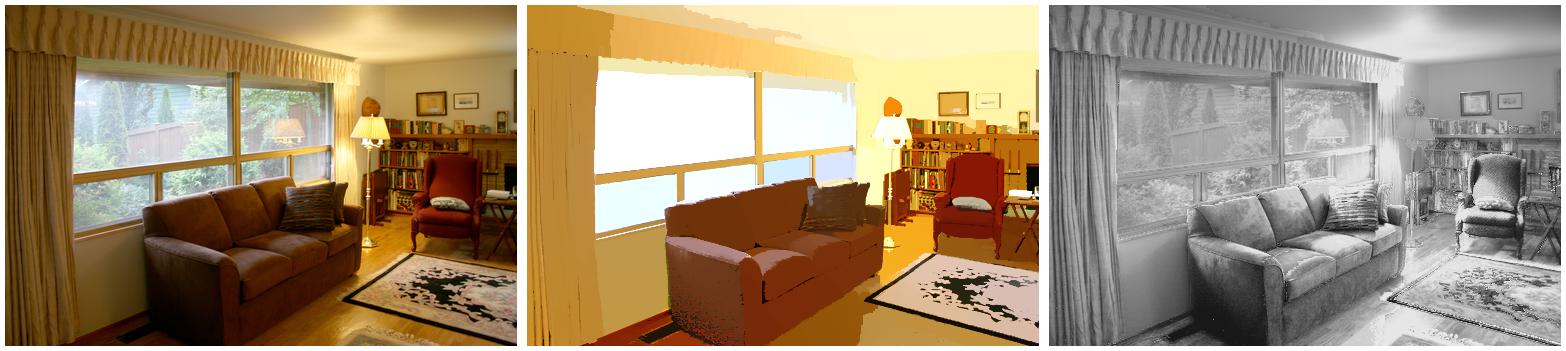} \\
  \includegraphics[width=\linewidth,height=2.4cm]{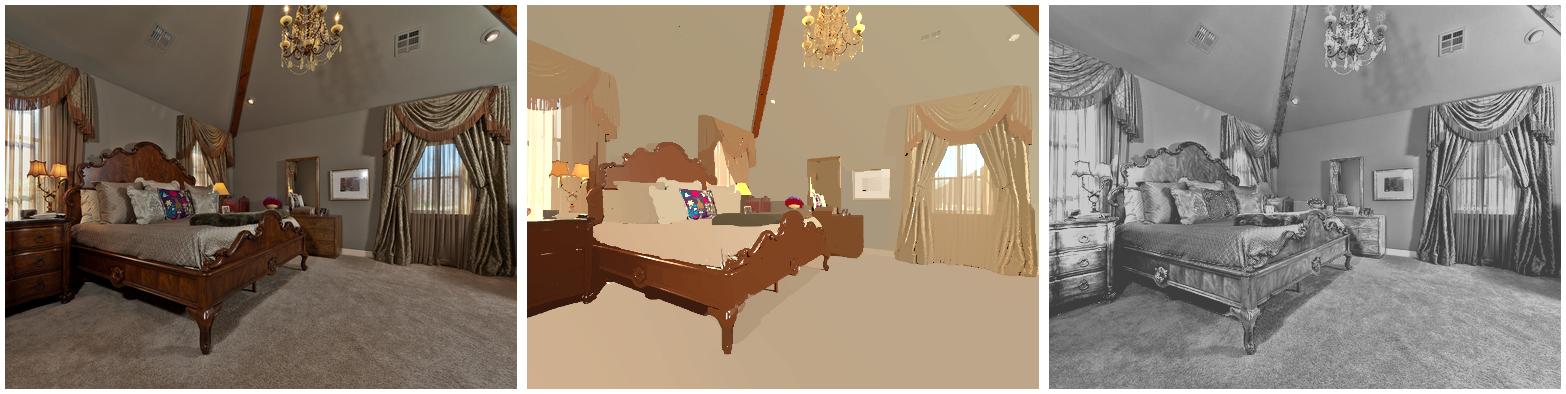}\\
  \includegraphics[width=\linewidth,height=2.4cm]{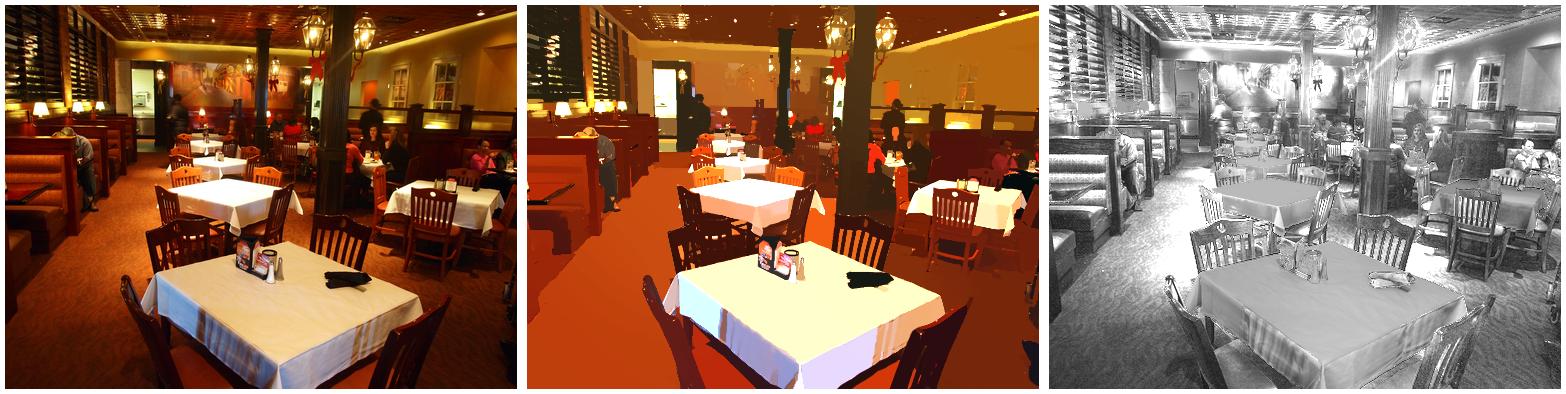} \\
  \includegraphics[width=\linewidth,height=2.4cm]{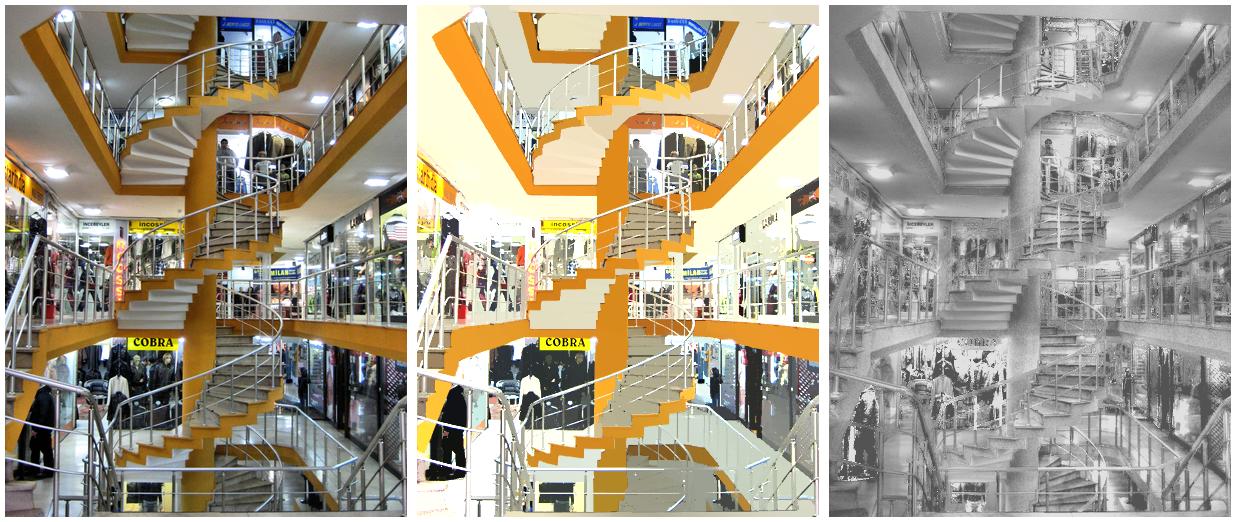} \\
  \includegraphics[width=\linewidth,height=2.4cm]{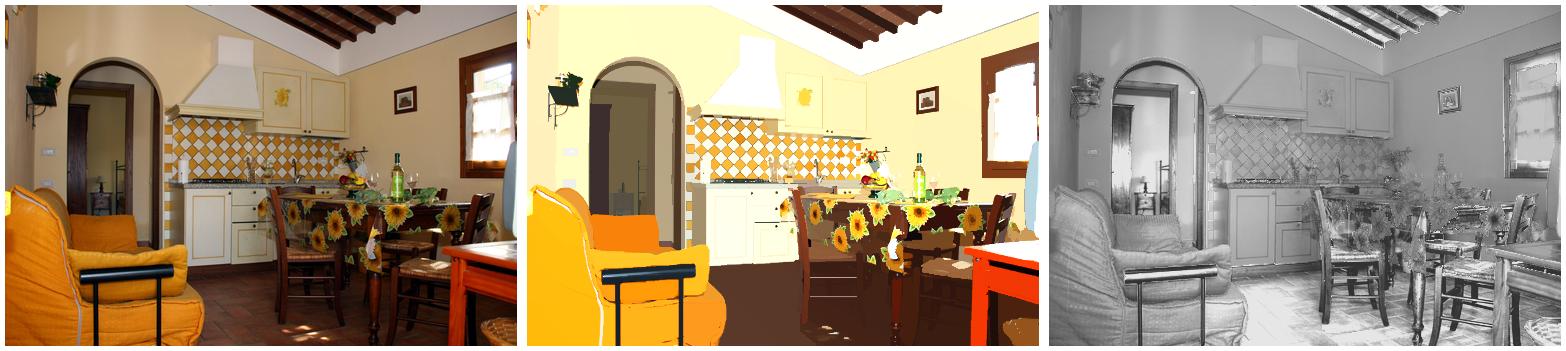} \\
    \end{minipage}
    \begin{minipage}{0.49\linewidth}
  \includegraphics[width=\linewidth,height=2.4cm]{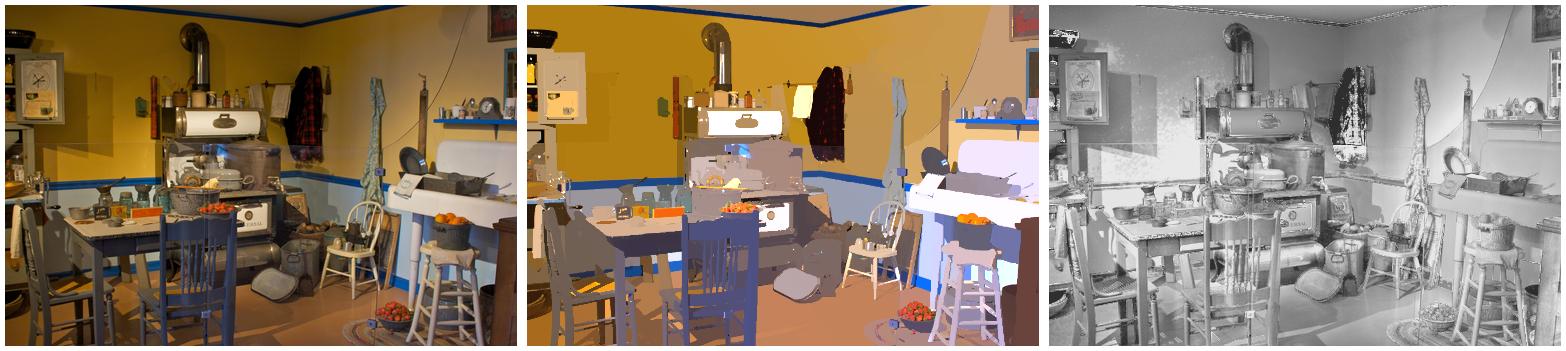} \\
  \includegraphics[width=\linewidth,height=2.4cm]{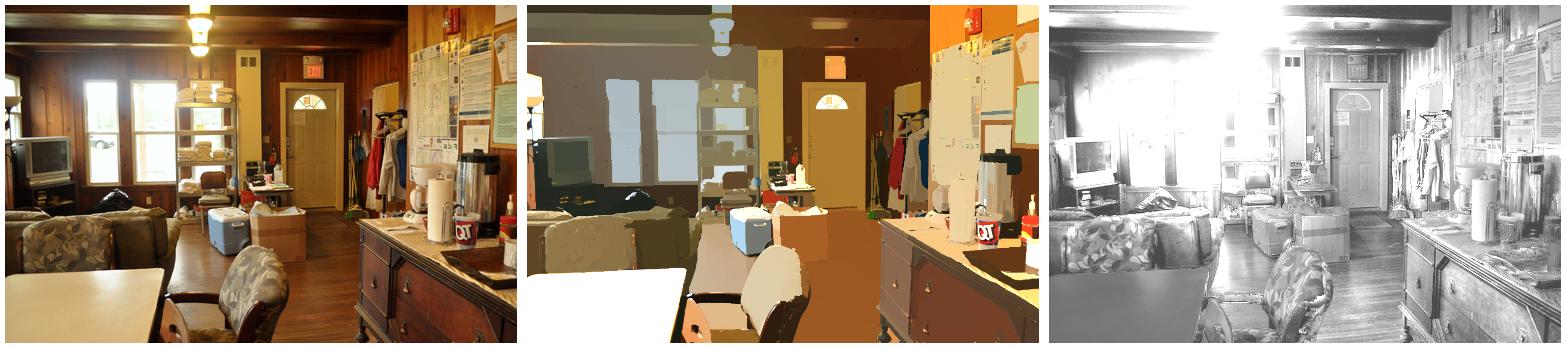} \\
  \includegraphics[width=\linewidth,height=2.4cm]{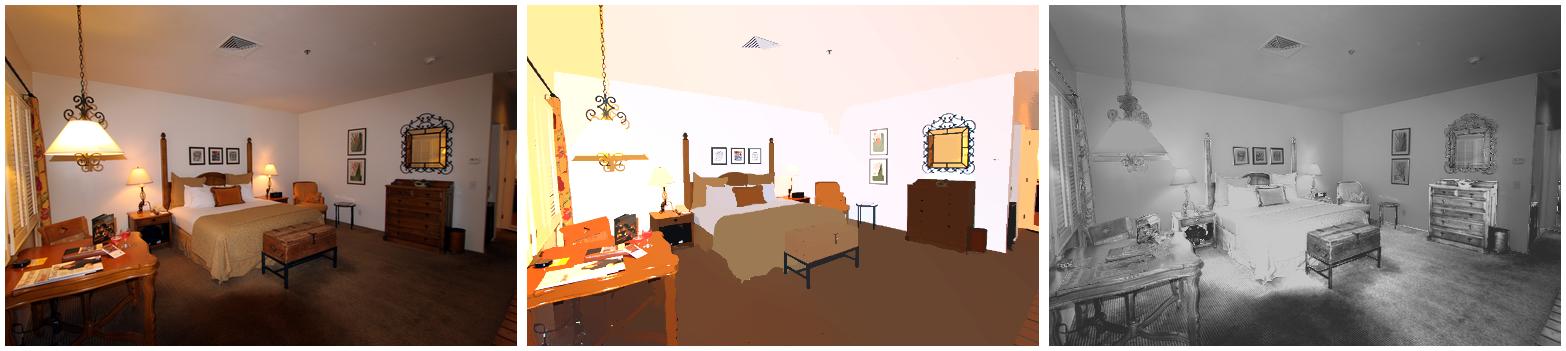} \\
    \includegraphics[width=\linewidth,height=2.4cm]{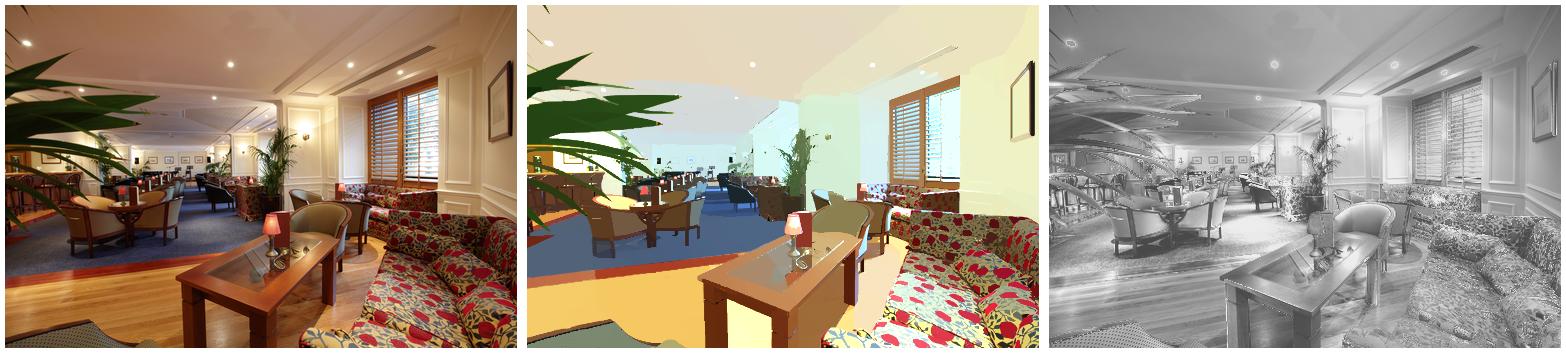} \\
  \includegraphics[width=\linewidth,height=2.4cm]{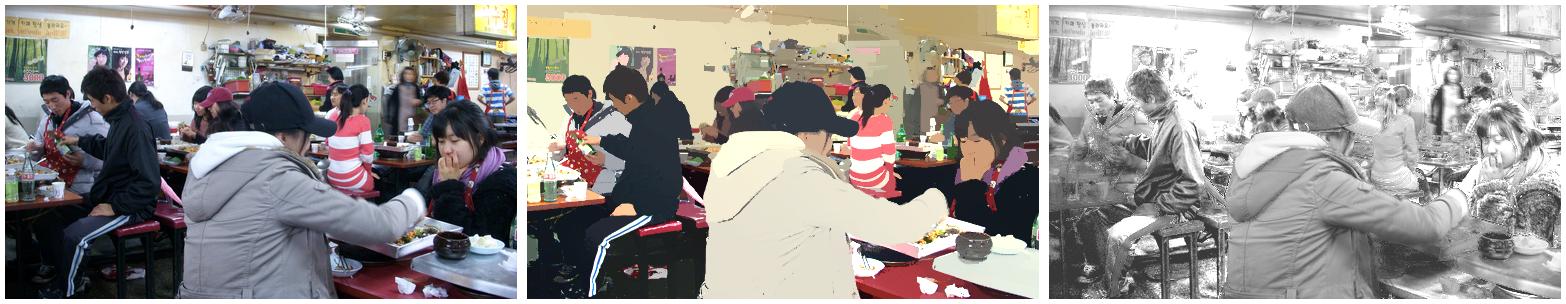} \\
  \includegraphics[width=\linewidth,height=2.4cm]{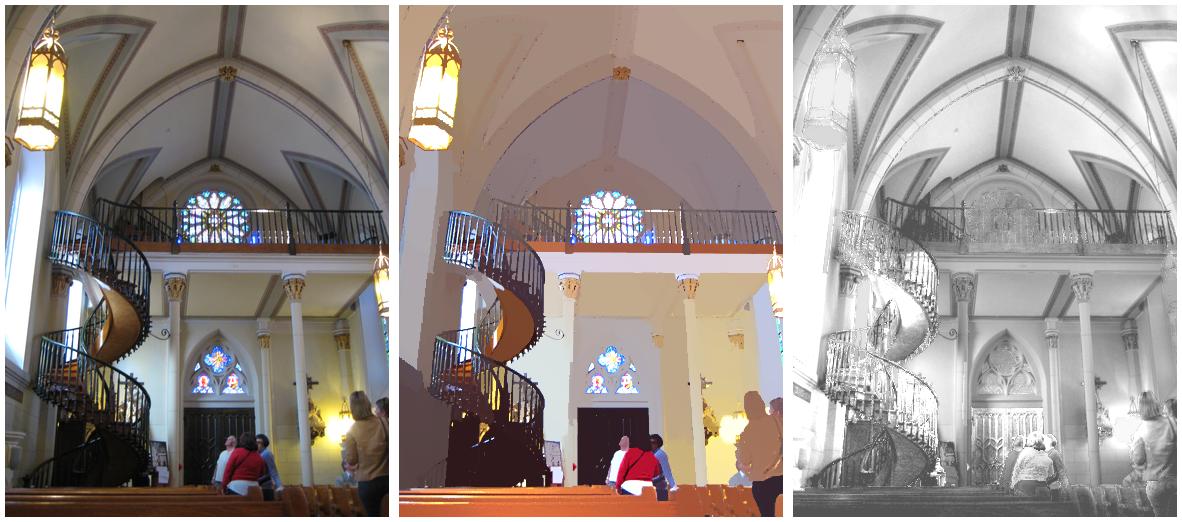} \\
    \end{minipage}
\caption{\textbf{Qualitative results:} (In each set from L to R) Original image, reflectance and shading on sample images from IIW dataset.
Notice separation of shadows and highlights in shading and colour sparsity in reflectance component.}
\label{fig:expMyRes}
\end{figure*}

\begin{figure}[h]
 \centering
 \includegraphics[width=\linewidth,height=9cm]{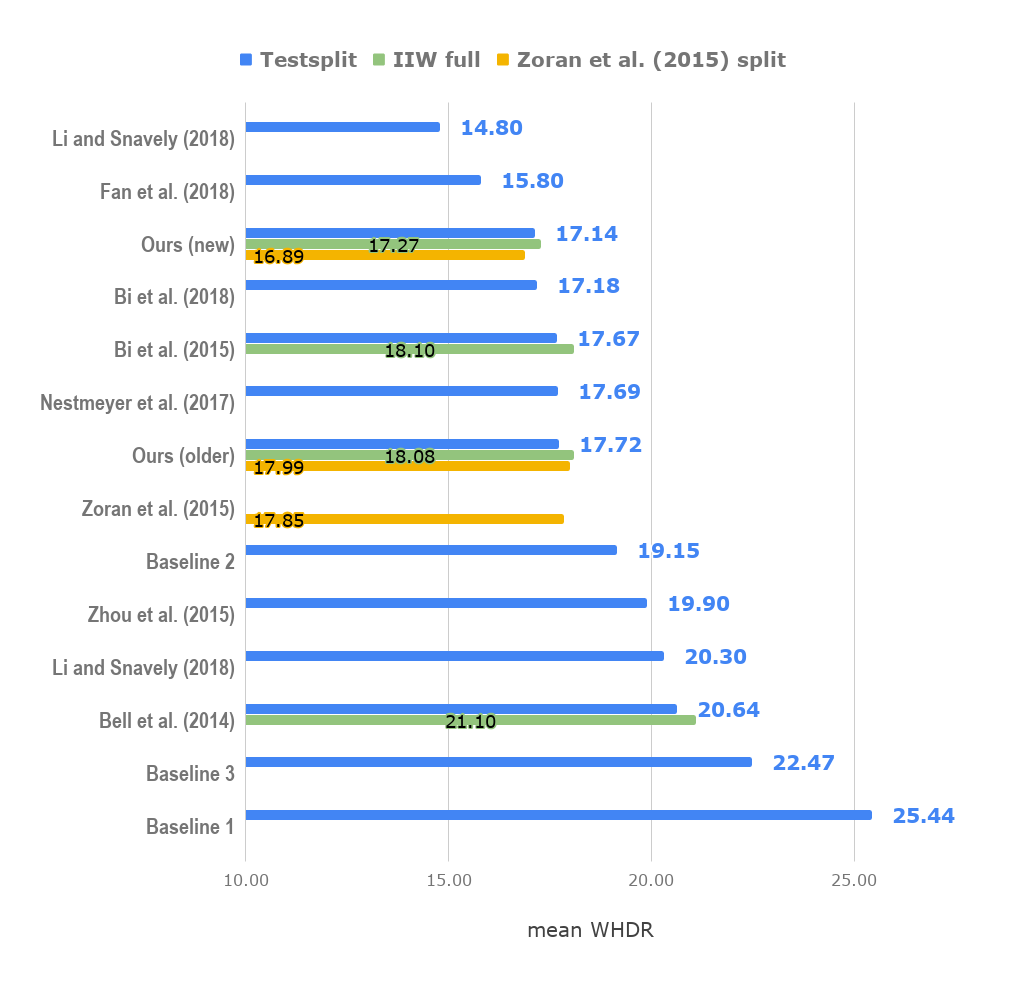}\\
 \caption{\textbf{Quantitative results:} Performance comparison between our method and other contemporary IID solutions.}
 \label{fig:whdrGraph}
\end{figure}

All our results are generated using a $5^{th}$ generation Intel i7 3.30 GHz desktop processor. 
Most of our prototype implementation is in Matlab with a few sections in C++ suggesting a significant scope of improving runtime efficiency. We present IID results from our method on a variety of datasets and evaluation measures as discussed below:

\paragraph{\textbf{IIW Dataset:}}
We show the results of our method on the IIW dataset in \fig \ref{fig:expMyRes} and \fig \ref{fig:whdrGraph}. Notice separation of shadows and illumination from light sources in the shading component and the colour consistency in the reflectance component.

\begin{figure*}[t!]
  \centering
  \begin{minipage}{\linewidth}
  \centering
    \includegraphics[width=0.92\linewidth,height=2.4cm]{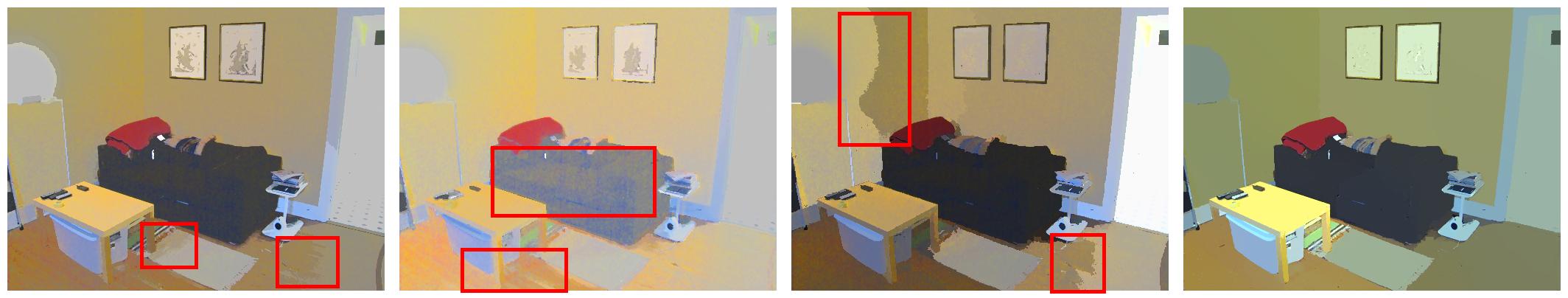} \\
    \includegraphics[width=0.92\linewidth,height=2.4cm]{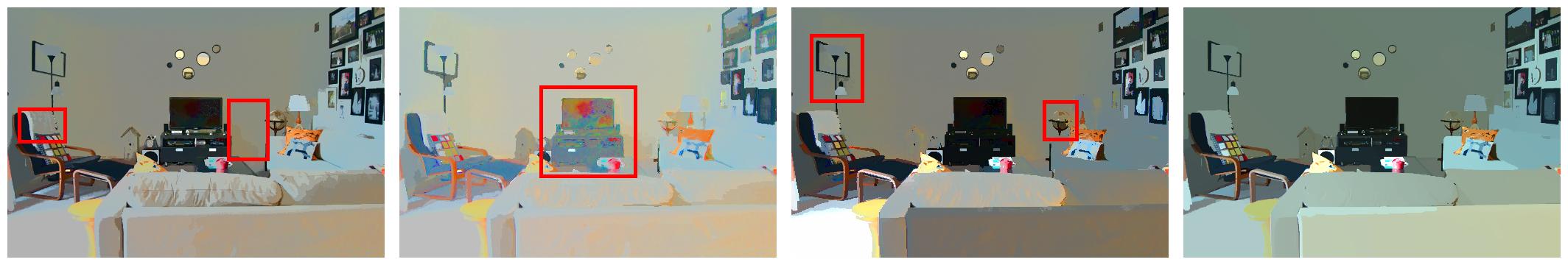} \\
    \includegraphics[width=0.92\linewidth,height=2.4cm]{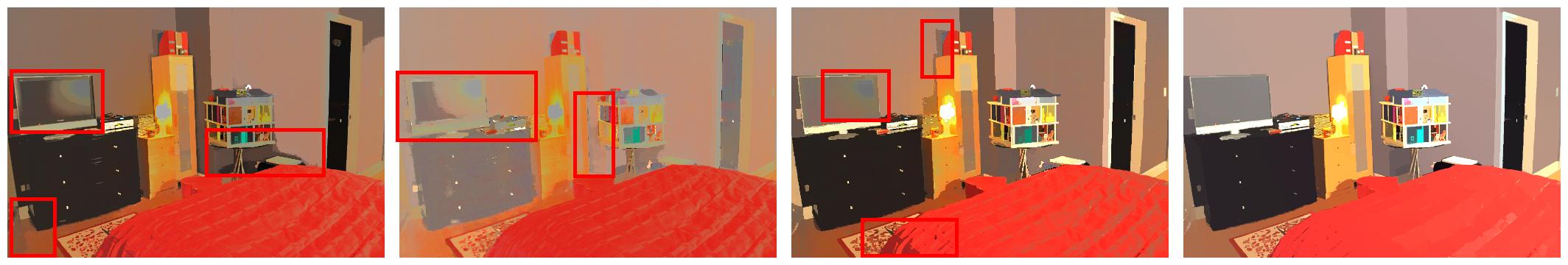} \\
  \end{minipage}
  \caption{\textbf{Qualitative comparisons:} (L to R) Reflectance from \cite{balaIID}, \cite{efrosIID}, \cite{biIID} and our method.
  Compared to the other methods shown, our framework produces results with fewer artifacts.}
  \label{fig:comparisonResults}
\end{figure*}

We compare our method quantitatively with other contemporary IID methods which encode scene information in terms of IID priors \citep{biIID, efrosIID, balaIID}.
The results are shown in \fig \ref{fig:whdrGraph} for the entire IIW dataset (green) and the test-split used in \cite{NarihiraSplit} (blue).
As \cite{efrosIID} use most of IIW dataset for training, we show their results only on the test-split.
The scores are reported as mentioned in the respective papers or downloaded from the respective project webpages.
We also compare our method with three baselines and on the test-split by \cite{freemanIID} (orange): 
\begin{itemize}
 \item \textit{Baseline 1}: only shading smoothness optimization.
 \item \textit{Baseline 2}: only reflectance sparsity optimization.
 \item \textit{Baseline 3}: edge preserving smoothing results from \cite{biIID} as reflectance.
\end{itemize}

Notice that our \textit{Baseline 2} performs better than both \cite{efrosIID} and \cite{balaIID} which highlights the strength of our reflectance priors.
\textit{Baseline 3} is computed directly from the edge-preserving smoothing results from \cite{biIID} and shows the difference of our framework from their underlying image flattening framework.
As can be seen from the graph in \fig \ref{fig:whdrGraph}, our method achieves significant error reduction in comparison to both \cite{balaIID} and \cite{efrosIID} on both the test-split and the full dataset (WHDR of $\mathbf{17.14}$ \vs $20.6$ and $19.9$ respectively).
Our method is competitive with both \cite{biIID} and \cite{filteringIID} (with WHDR $17.67$ and $17.69$ respectively) but with lesser artifacts in reflectance results (\fig \ref{fig:comparisonResults}).
Additional comparisons with previous IID methods like \cite{zhaoIID} and \cite{clusteringIID}, with WHDR as $23.20$ and $25.46$ respectively (are not shown in graph for the sake of clarity).
The error could be further reduced if we allow for manual tuning of $k$ parameter for each image, chosen based on image complexity (textures, colours, lighting \etc). 
Note that in our method is more direct as there is no need to perform separate clustering, classification or CRF labeling steps.
Our semantic priors lead to consistent reflectance values with lesser number of patchy artifacts. Furthermore our approach is better at handling chromatic noise as can be seen in the reflectance of dark regions in the results. 

Parallel to our work in this paper, there are a few recent direct deep learning solutions by \citet{newBiIID}, \citet{msrIID} and two works from \citet{snavelyIID_ECCV,snavelyIID}. The respective WHDR scores on the test-split are $17.18$, $15.80$, $20.3$ and $14.80$.
\cite{snavelyIID} and \cite{newBiIID} introduce new datasets for training.
They use the illumination invariant property of reflectance from time lapse videos or synthetically rendered scenes as a prior for IID. 
\cite{msrIID} take inspiration from \cite{filteringIID} and perform guidance filtering within the CNN framework rather than a separate post processing step 
which leads to significant error reduction. 
Based on this observation, we think that properly incorporating semantic information (perhaps in the form of region proposals or masks) within the deep network architecture would further improve the IID performance of such networks. 

\begin{table}[t!]
\centering
\caption{Results on ARAP dataset by \cite{IIDstarreport} using LMSE metrics (lower is better).}
\begin{minipage}{\linewidth}
    \centering
  \begin{adjustbox}{max width=0.98\textwidth}
    \begin{tabular}{c | c | c || c }
      \multicolumn{4}{c}{\textbf{Results on synthetic images}}  \\
      \hline
      Image 	& \multicolumn{3}{c}{Our Method} \\
      title 	& R & S & Mean \\
      \hline
      \hline
      arabic 	& 0.0158 & 0.0138 & 0.0148 \\
      babylone 	& 0.0036 & 0.0055 & 0.0046 \\
      breakfast & 0.0047 & 0.0083 & 0.0065 \\
      head 	& 0.0122 & 0.0275 & 0.0198 \\
      italian 	& 0.0163 & 0.0258 & 0.0210 \\
      san miguel& 0.0277 & 0.0251 & 0.0264 \\
      sponza	& 0.0482 & 0.0108 & 0.0295 \\
      villa 	& 0.0331 & 0.0743 & 0.0537 \\
      whiteroom & 0.0093 & 0.0189 & 0.0141 \\
      \hline
      \textbf{Average} 	& 0.0190 & 0.0233 & \textbf{0.0212} \\
    \end{tabular}
  \end{adjustbox}
\end{minipage}
\label{tab:arap}
\end{table} 

\paragraph{\textbf{ARAP Dataset:}} For assessing IID performance \cite{IIDstarreport} presented several realistic synthetically rendered images of various sizes and content. We show LMSE performance of our method on these images in \tab \ref{tab:arap} for both the components. As reported on the authors' project webpage\protect\footnote{\protect\url{https://perso.liris.cnrs.fr/nicolas.bonneel/intrinsicstar}}, average LMSE for these images for \cite{balaIID}, \cite{efrosIID}, \cite{barronIID_PAMI}, \cite{narihiraIID}, \cite{bonneelIID} (automatic) and with scribbles are $0.0262$, $0.0191$, $0.0352$, $0.0212$, $0.0240$ and $0.0215$ respectively.
While other results have been reported after tuning parameters to their most suitable values for this dataset, we present our results using the consistent set of values as presented in \sect \ref{sec:experiments}.
Still our average performance is quite competitive on the standard metrics. Further improvement could be achieved by tuning the system parameters (\eg simply changing $k=3$ reduces mean error value to $0.0204$).

\begin{table}[t!]
\centering
\caption{Average Precision percentage results on SAW test-split from \cite{sawIID} (higher is better).}
\begin{minipage}{\linewidth}
    \centering
  \begin{adjustbox}{max width=0.98\textwidth}
    \begin{tabular}{c | c | c | c | c | c | c | c}
      \multicolumn{8}{c}{\textbf{Shading component evaluation on SAW dataset}}  \\
      \hline
      Iterations & 1 & 2 & 3 & 4 & 5 & 6 & 7 \\ 
      \hline
      Stage 1 & 96.57 & 96.73 & 96.96 & 97.24 & \textbf{97.59} & 98.01 & 98.26 \\
      \hline
      Stage 2 & 88.93 & 88.53 & 88.75 & 89.87 & \textbf{91.54} & 93.50 & 95.68 \\
      \hline      
    \end{tabular}
  \end{adjustbox}
\end{minipage}
\label{tab:saw}
\end{table} 

\paragraph{\textbf{SAW Dataset:}}
As WHDR is designed to evaluate only the reflectance decomposition performance of an IID method, \cite{sawIID} extended IIW dataset with manually labeled shading ground truth regions which could be used for shading component evaluation. We present the results (average precision percentage) over successive iterations from both stages of our method in \tab \ref{tab:saw} using the standard test-split of 1699 images as provided in \cite{sawIID}. As our system focuses separately on shading and reflectance in two stages of the optimization, the results of our \textit{Stage 1} here outperforms \textit{Stage 2} outputs. Furthermore after initial setup, performance improves for both the stages over successive iterations proving validating the iterative procedure. For comparison AP $\%$ scores for \cite{balaIID}, \cite{efrosIID} and \cite{filteringIID} are $97.37$, $96.24$ and $96.85$ respectively.
Although later iterations have better performance scores but they are not qualitatively superior due to increasing smoothing of \textit{Stage 1} shading and sparsity of \textit{Stage 2} reflectance. This also highlights a major challenge with quantitative evaluations of IID components as current metrics like WHDR, AP $\%$ or LMSE, do not reflect the perceptual quality and applicative utility of the results. As suggested by \cite{IIDstarreport} a better method of evaluating IID methods is by using them in image editing scenarios which is the end goal of IID in computer vision research. We present such evaluations in the next section \sect \ref{sec:applications}.

\begin{figure*}
\centering
    \begin{minipage}{0.49\linewidth}
    \centering
      \includegraphics[width=\linewidth,height=2.4cm]{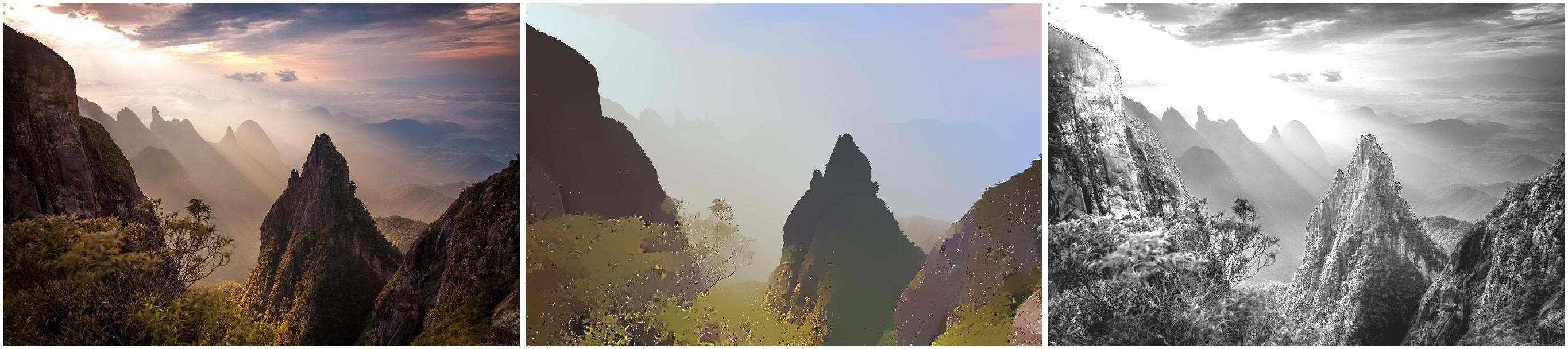} \\
      \includegraphics[width=\linewidth,height=2.4cm]{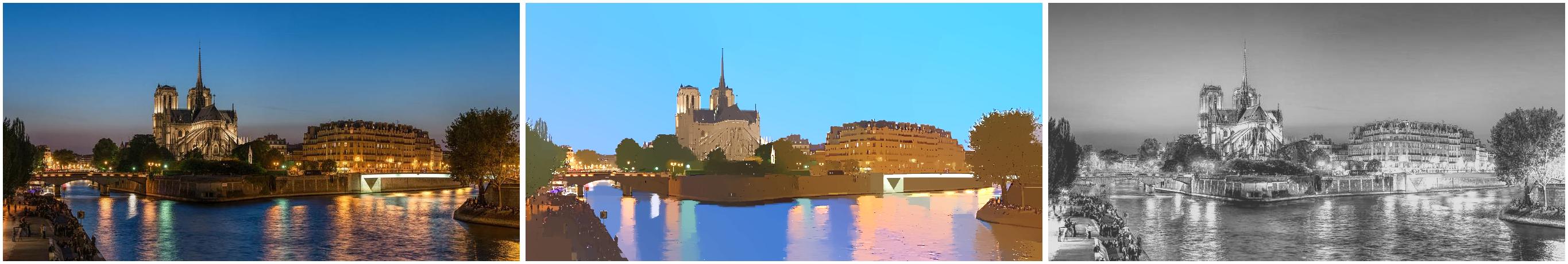} \\
      \includegraphics[width=\linewidth,height=2.4cm]{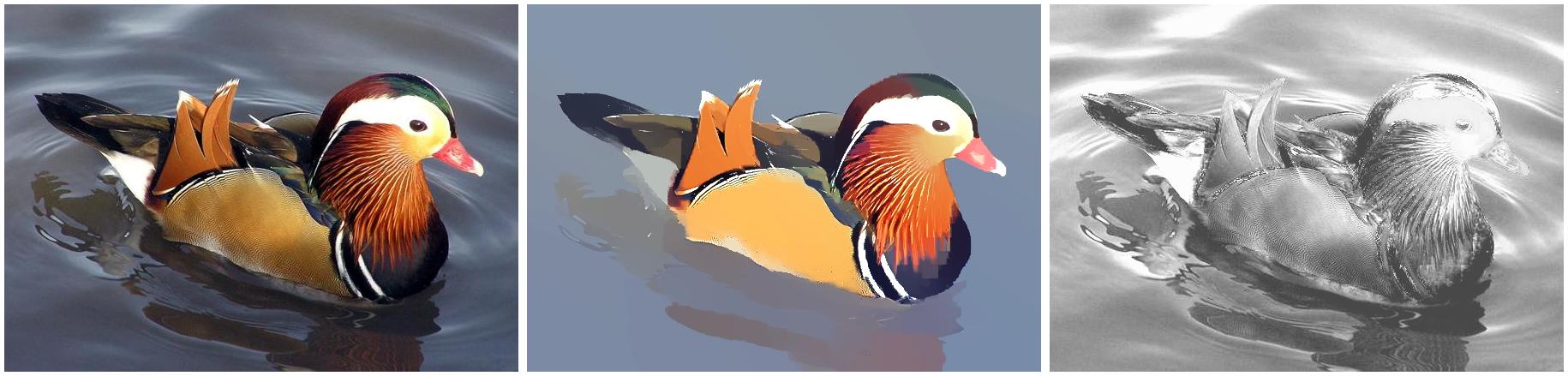} \\
      \includegraphics[width=\linewidth,height=2.4cm]{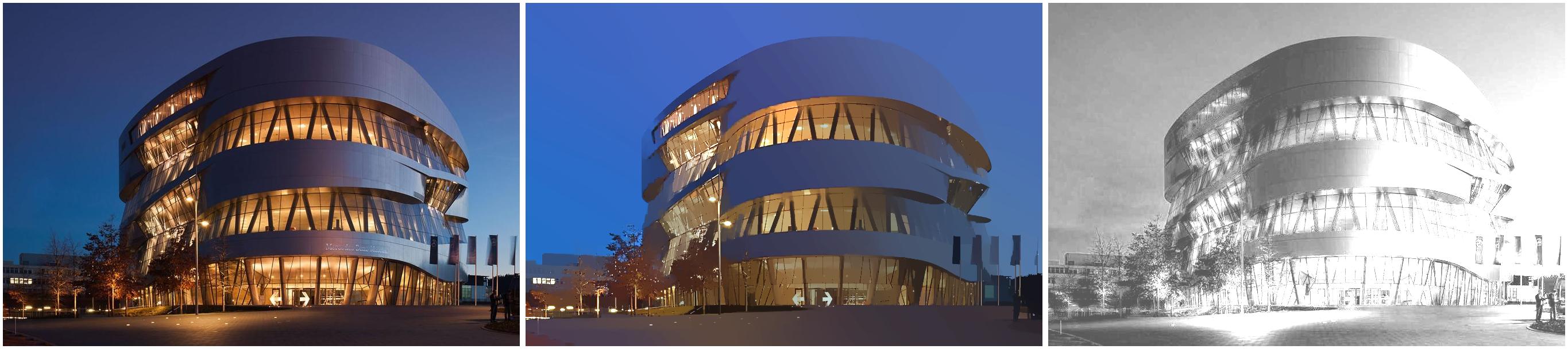} \\
    \end{minipage}
    \begin{minipage}{0.49\linewidth}
    \centering
      \includegraphics[width=\linewidth,height=2.4cm]{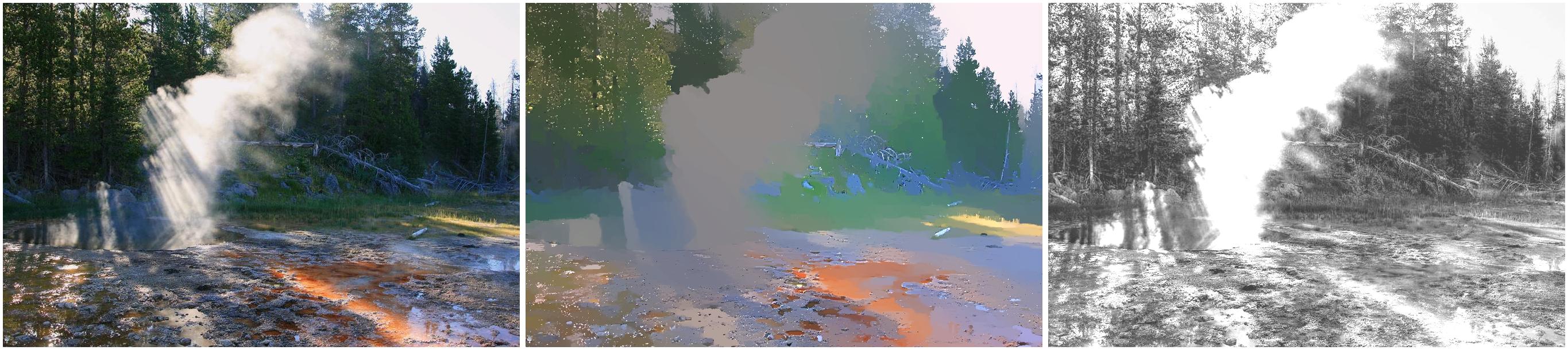} \\
      \includegraphics[width=\linewidth,height=2.4cm]{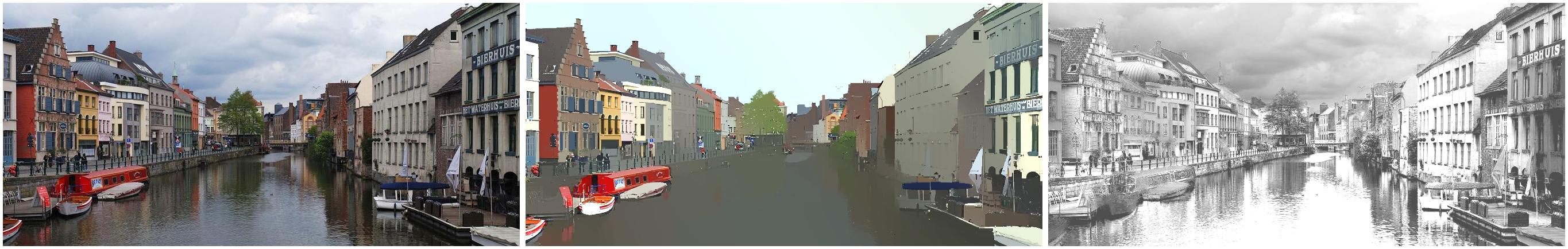}\\
      \includegraphics[width=\linewidth,height=2.4cm]{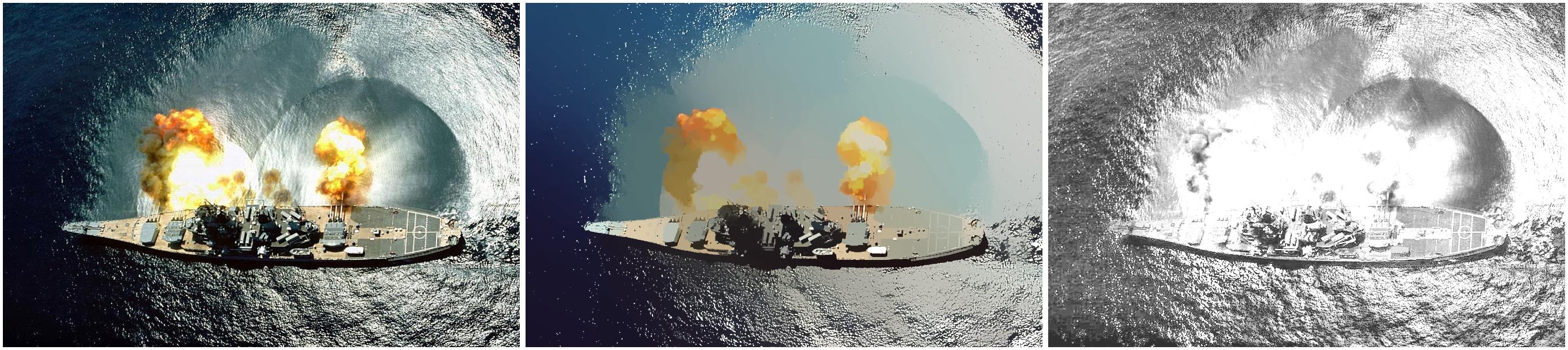} \\
      \includegraphics[width=\linewidth,height=2.4cm]{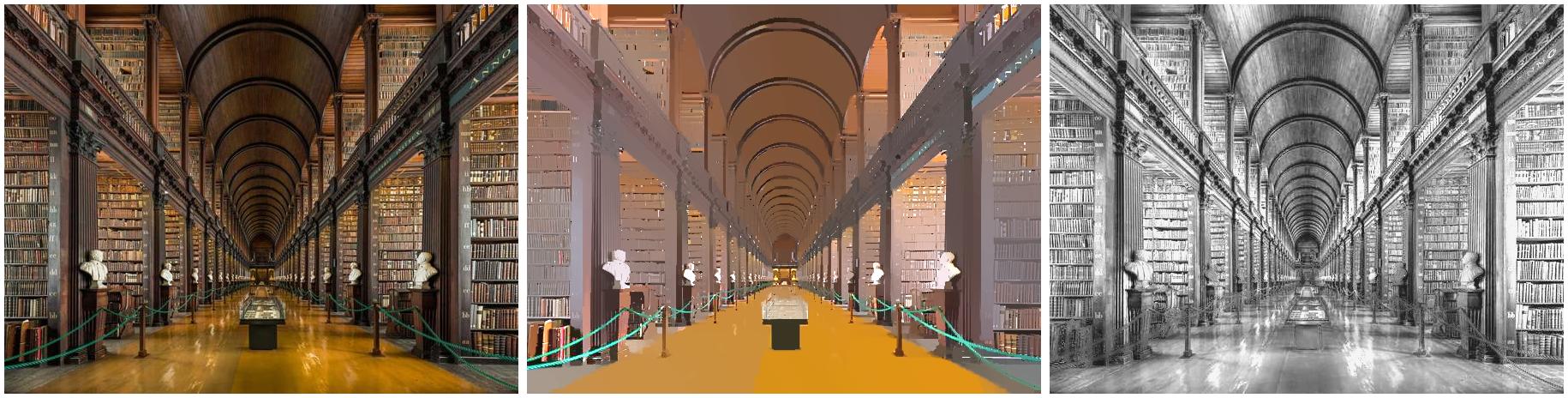} \\
    \end{minipage}
    \caption{Our results on challenging Internet images highlighting generality of our method for a variety of scene types, scene complexity and lighting. The scenes shown contain variations like indoors/outdoors, artificial/natural lighting, day/night setting, complex/simple subjects, multiple/single light sources and dynamic/static.}
      \label{fig:internetResults}
\end{figure*}

\paragraph{\textbf{Wikimedia Dataset:}}
To explore the generality of our method beyond manually curated evaluation datasets (which mostly comprise of indoor scenes or a few synthetic images), we also experimented with diverse and challenging publicly available images\protect\footnote{\protect\url{https://www.wikimedia.org}} directly downloaded from the Internet. As shown in \fig \ref{fig:internetResults}, our method works on several scene types (indoor, outdoors, natural, cityscapes \etc) with varying complexity (single object \vs multiple objects) and diverse lighting configurations (single \vs multiple light sources, natural \vs artificial lighting, day \vs night lighting \etc) generating plausible decompositions.
 
\begin{figure*}[t!]
  \centering
  \begin{minipage}{0.49\linewidth}
  \centering
    \includegraphics[width=0.98\linewidth,height=2.8cm]{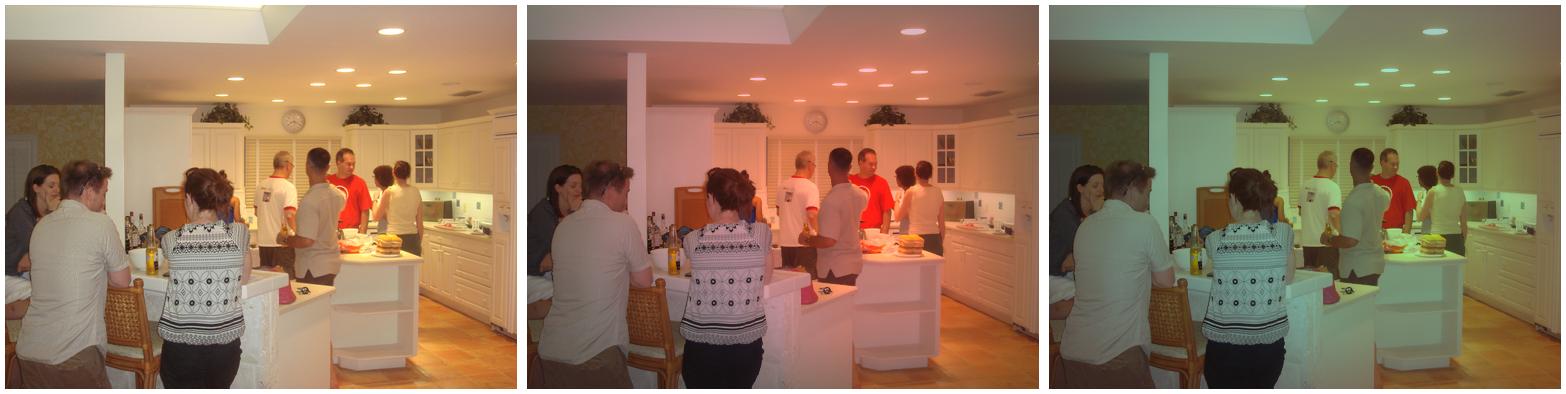} \\
    \includegraphics[width=0.98\linewidth,height=2.8cm]{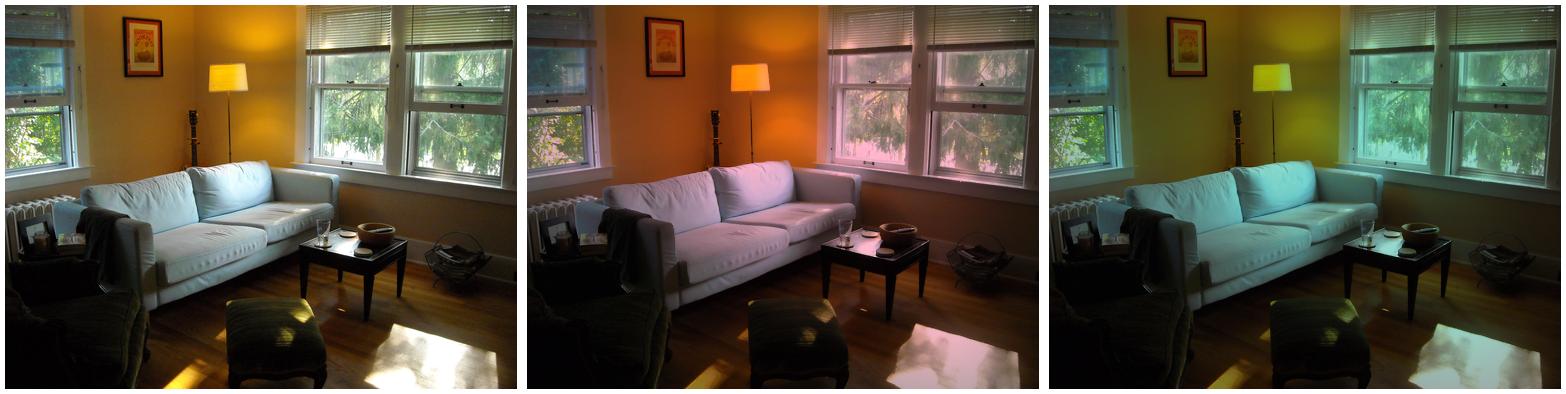} \\
    \includegraphics[width=0.98\linewidth,height=2.8cm]{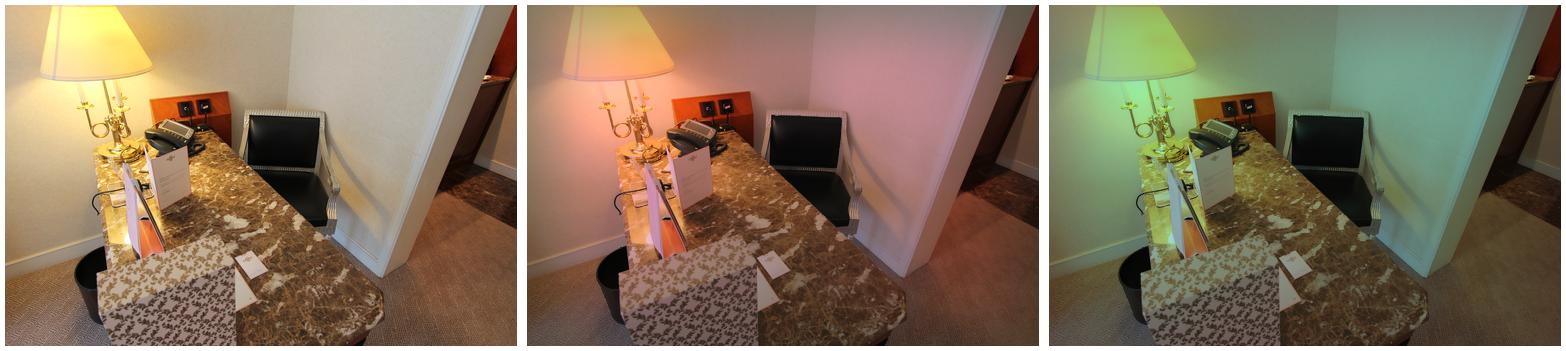} \\
    \includegraphics[width=0.98\linewidth,height=2.8cm]{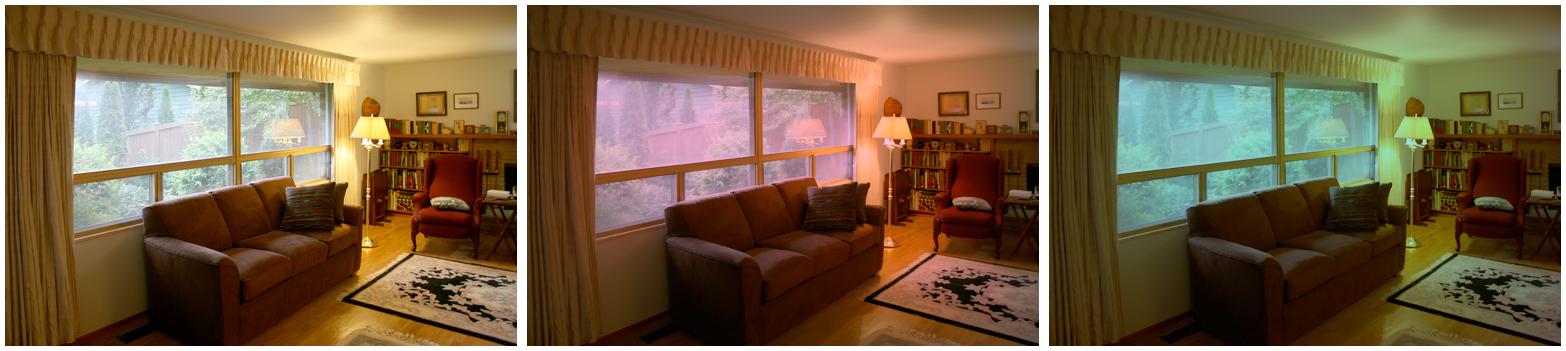} \\
  \end{minipage}
  \begin{minipage}{0.49\linewidth}
  \centering
    \includegraphics[width=0.98\linewidth,height=2.8cm]{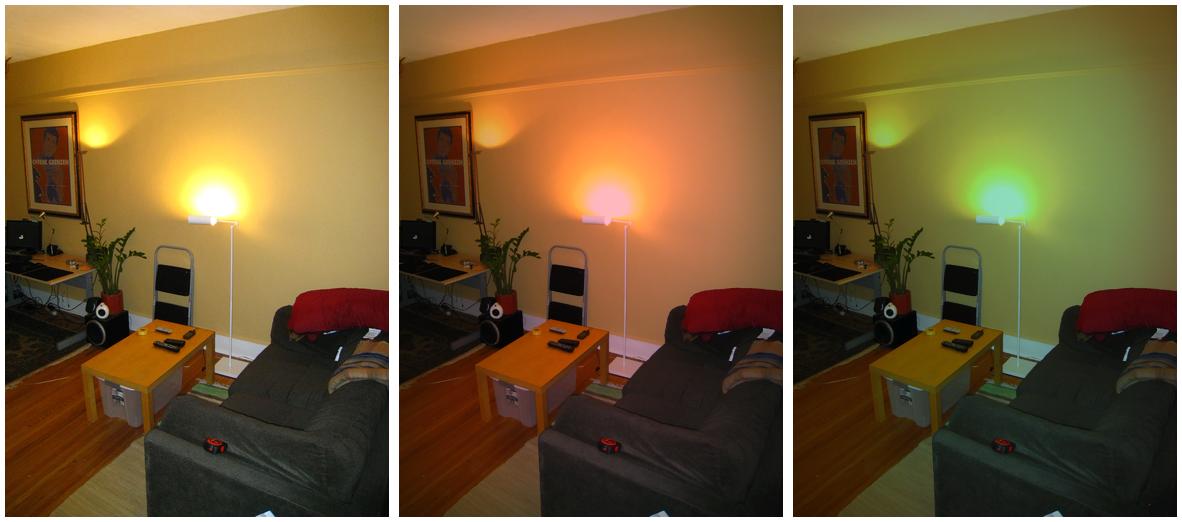} \\
    \includegraphics[width=0.98\linewidth,height=2.8cm]{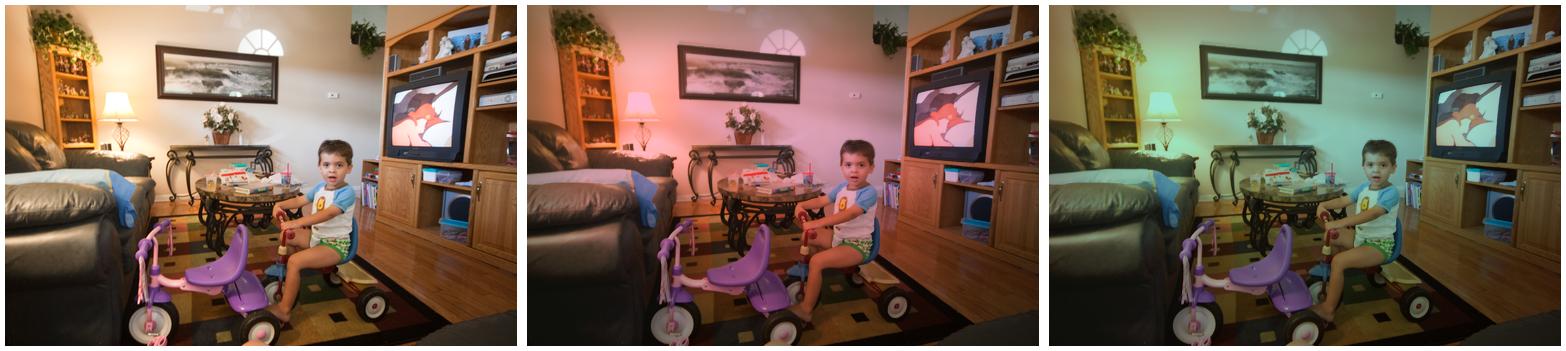} \\
    \includegraphics[width=0.98\linewidth,height=2.8cm]{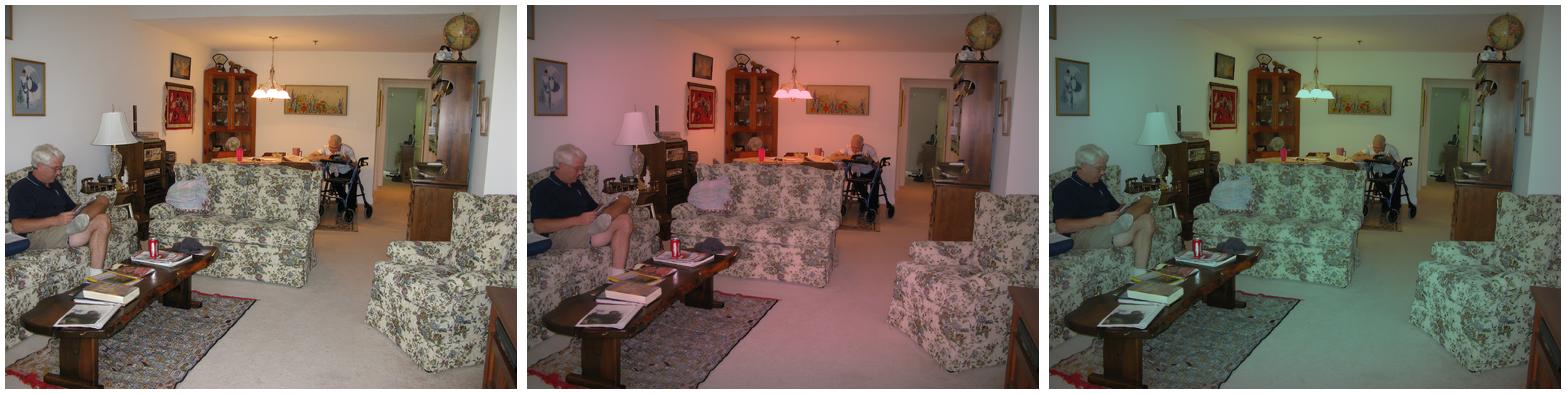} \\
    \includegraphics[width=0.98\linewidth,height=2.8cm]{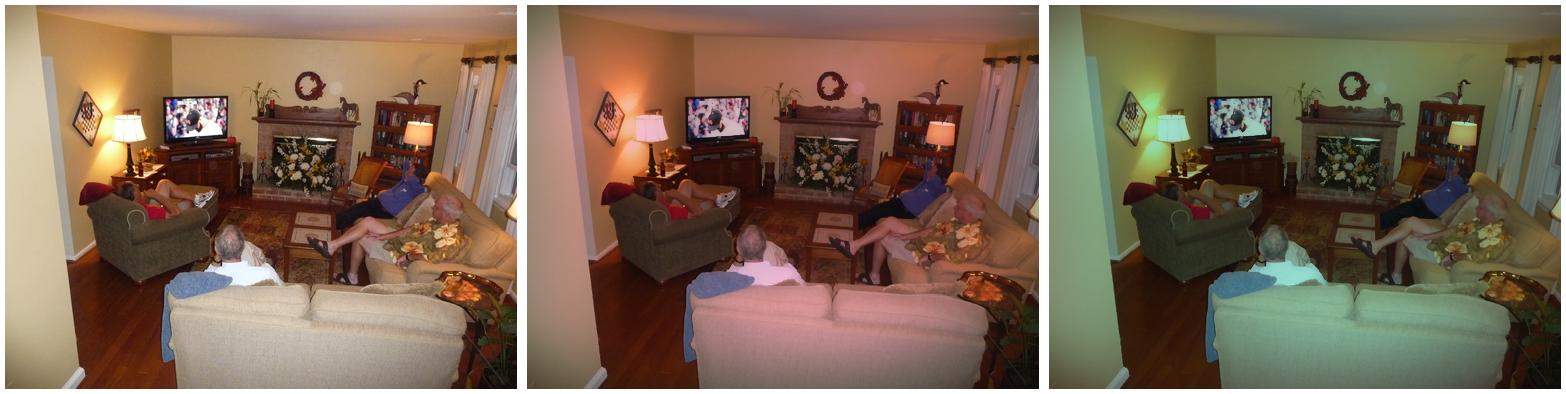} \\
  \end{minipage}
  \caption{\textbf{Illumination colour manipulation:} Using our approximated lighting region and colour estimates, we can modify the illumination colour in the scene, thereby altering the tone of the image. Here each scene is shown in original, red illumination and green illumination respectively. }
  \label{fig:appResults_2}
\end{figure*}

\begin{figure*}[t!]
  \centering
  \begin{minipage}{0.49\linewidth}
  \centering
    \includegraphics[width=0.92\linewidth,height=3cm]{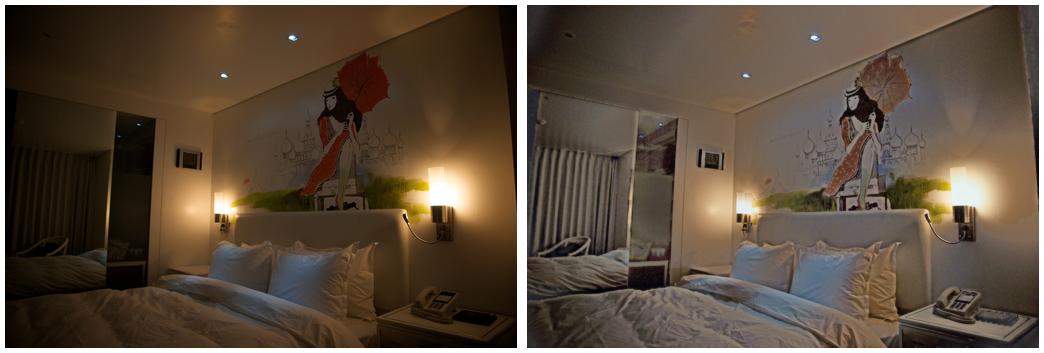} \\
    \includegraphics[width=0.92\linewidth,height=3cm]{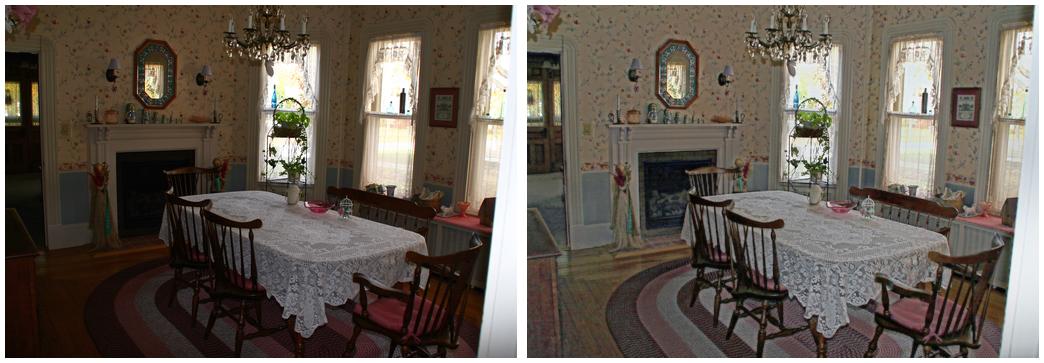} \\
    \includegraphics[width=0.92\linewidth,height=3cm]{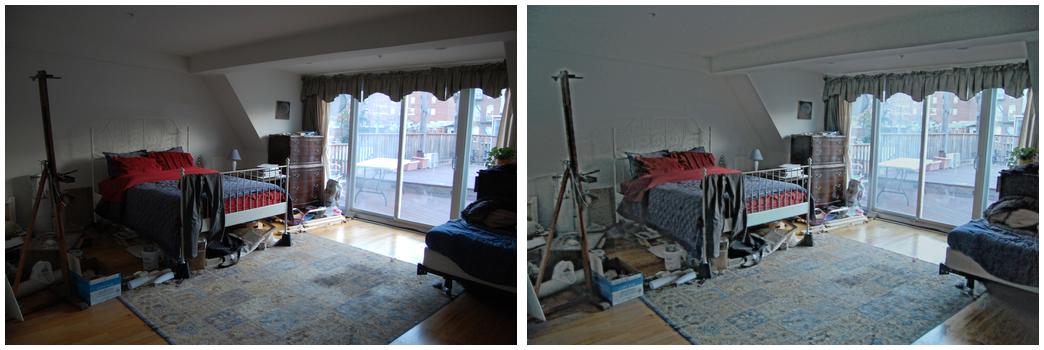} \\
    \includegraphics[width=0.92\linewidth,height=3cm]{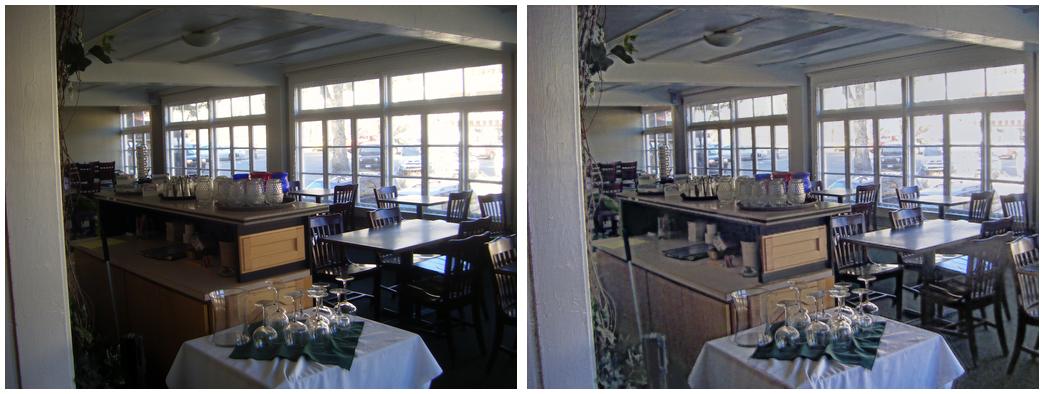} \\
  \end{minipage}
  \begin{minipage}{0.49\linewidth}
  \centering
    \includegraphics[width=0.92\linewidth,height=3cm]{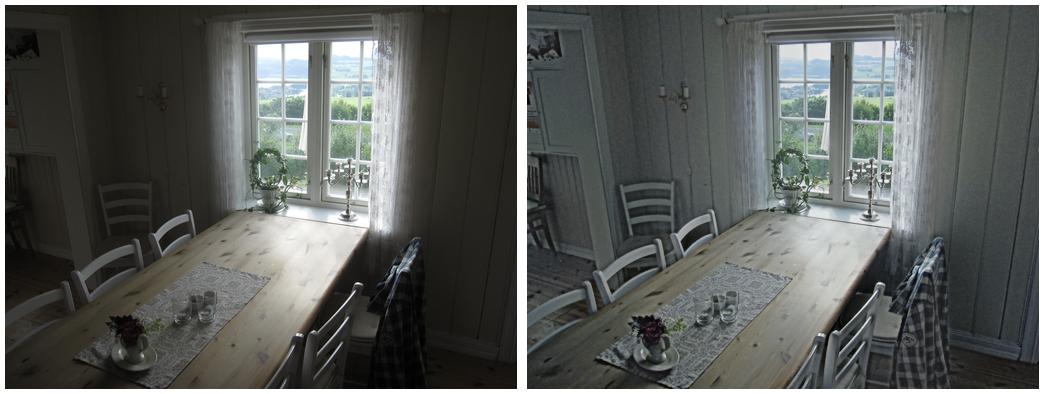} \\
    \includegraphics[width=0.92\linewidth,height=3cm]{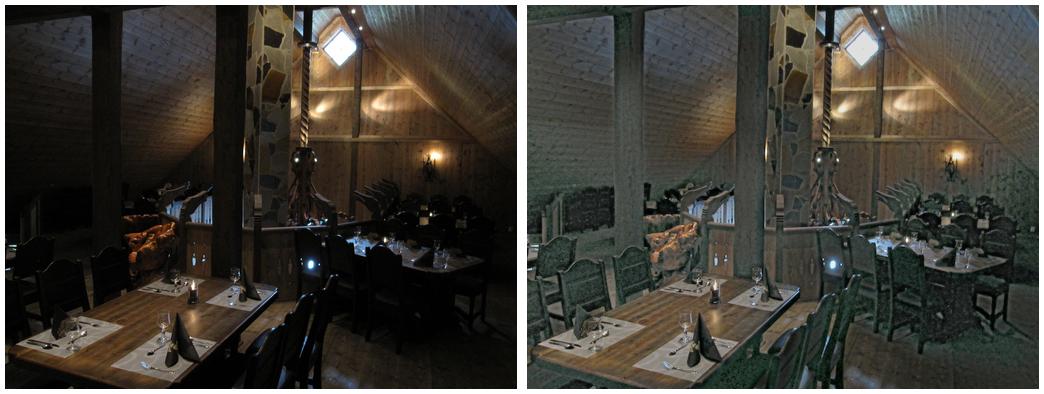} \\
    \includegraphics[width=0.92\linewidth,height=3cm]{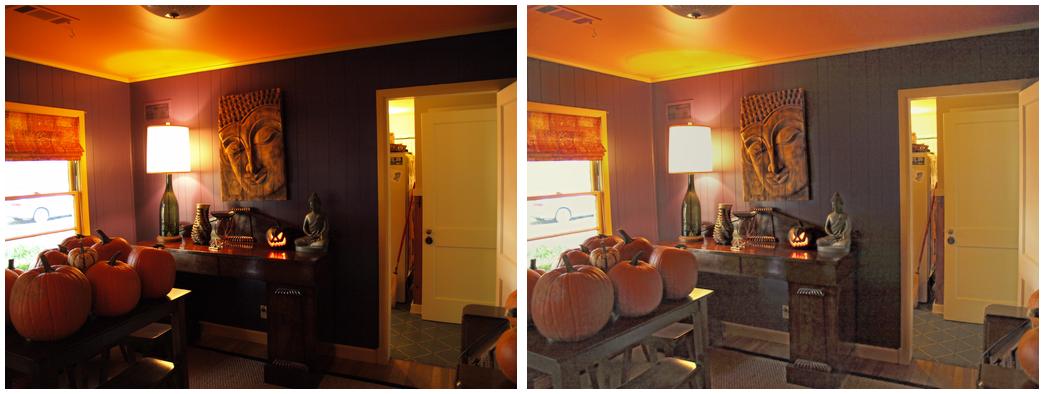} \\
    \includegraphics[width=0.92\linewidth,height=3cm]{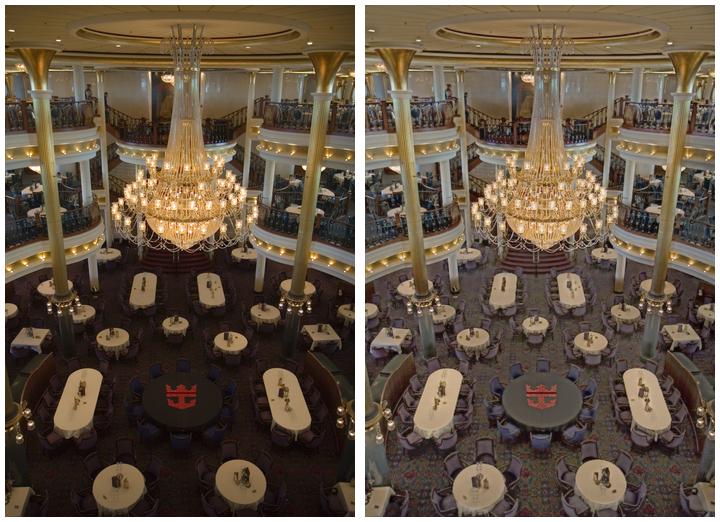} \\
  \end{minipage}
  \caption{\textbf{Illumination intensity manipulation:} We increase the contribution of our approximated indirect light component without altering the direct component to enhance the visibility of the dark regions in the image. Here for each scene original image and the relit versions are shown. }
  \label{fig:appResults_1}
\end{figure*}


\section{Applications}
\label{sec:applications}

In this section we employ the results obtained from the two stages of our framework to present image editing based evaluation and additional two novel IID applications.
As the two stages are designed to separately optimize smoothness and sparsity, we merge the results from both the stages, \ie $\rho$ and $\sigma$ from the shading optimization and similarly $R$ and $S$ from the reflectance optimization, to harness both the properties.
We cross multiply the components from the two stages to obtain a sparse ($I_{R\sigma} = R \cdot \sigma$) and a detailed ($I_{\rho S} = \rho \cdot S$) reconstruction of the original image ($I$). 
We then find the fractional residues $C_1$ and $C_2$ from these two reconstructions as:
\[
 C_1 = \frac{I}{I_{R\sigma}} \quad \text{and} \quad
 C_2 = \frac{I}{I_{\rho S}} 
\]
and take the Gaussian filtered mean of these two estimates to obtain our illumination colour approximation $C$. Now we can update our corresponding shading components by dividing $C$ from $S$ and multiplying it to $\sigma$ and taking the mean of both the results to obtain the merged shading and reflectance components.
\begin{figure*}[t!]
  \centering
  \begin{minipage}{0.49\linewidth}
  \centering
    \includegraphics[width=1\linewidth,height=2.2cm]{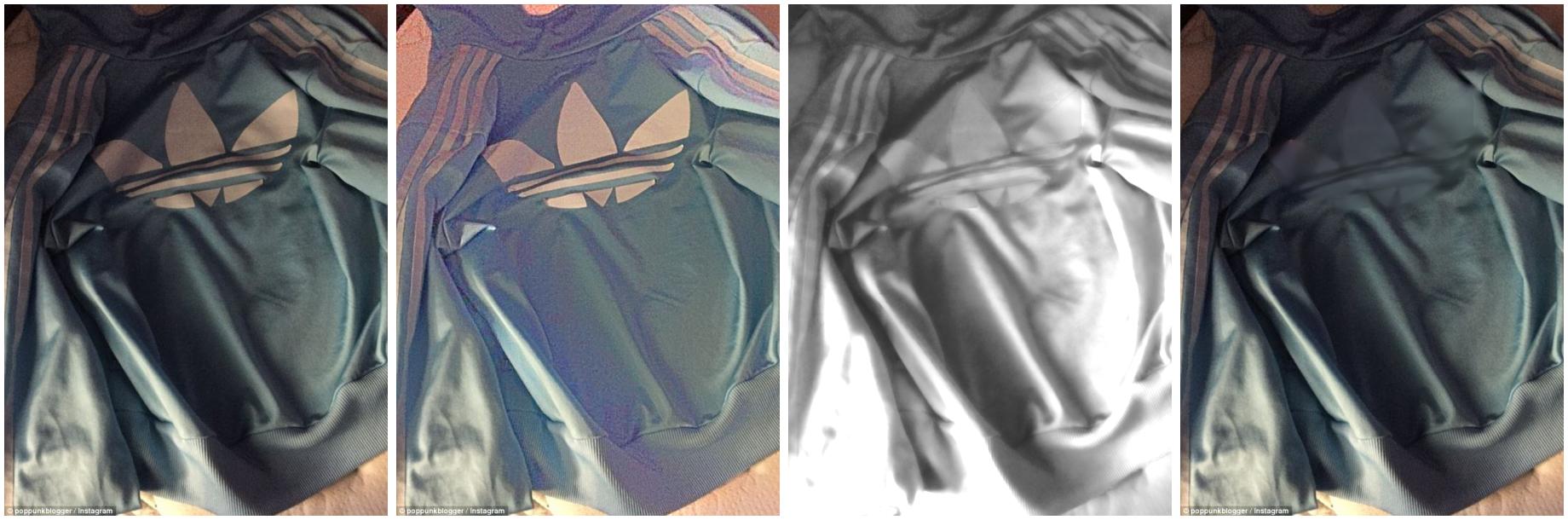} \\
    \includegraphics[width=1\linewidth,height=2.2cm]{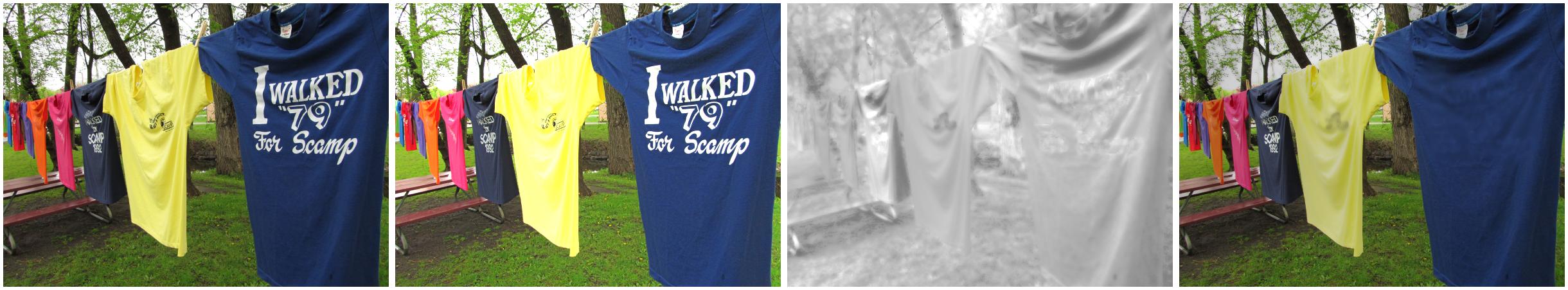} \\
    \includegraphics[width=1\linewidth,height=2.2cm]{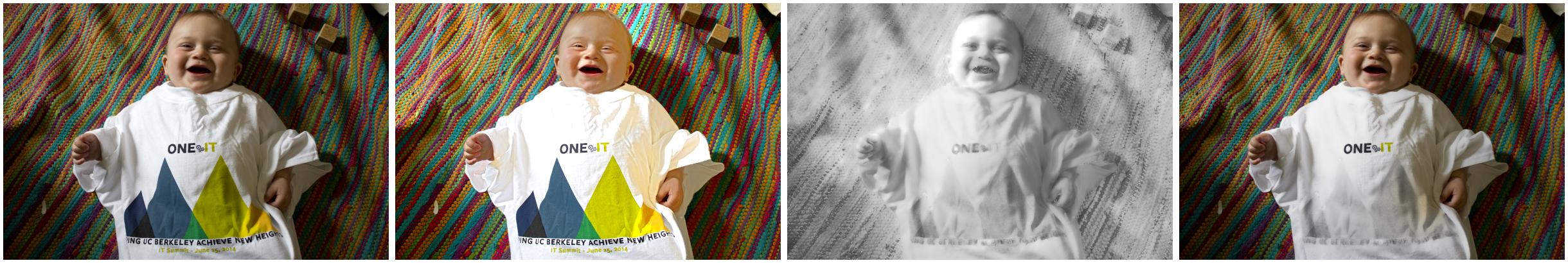} \\
    \includegraphics[width=1\linewidth,height=2.2cm]{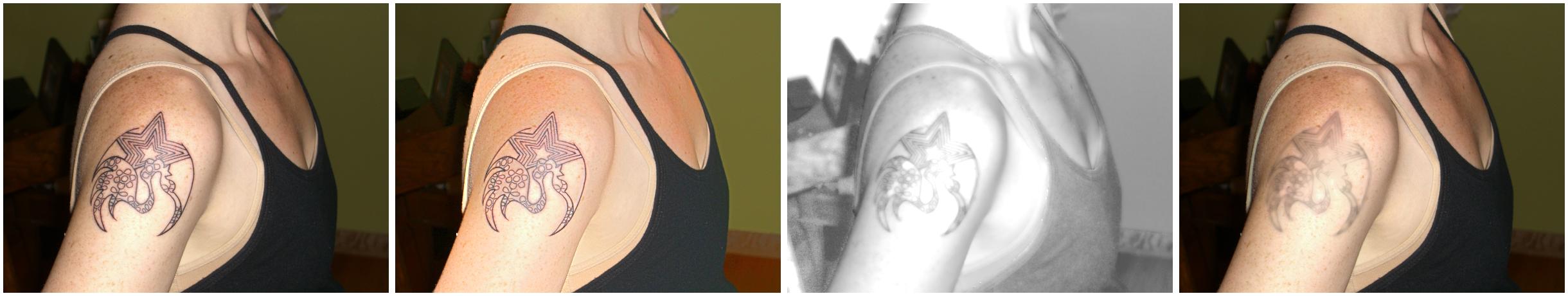} \\
    \rule{1\linewidth}{0.5pt}\vspace{2mm}
    \includegraphics[width=1\linewidth,height=2.2cm]{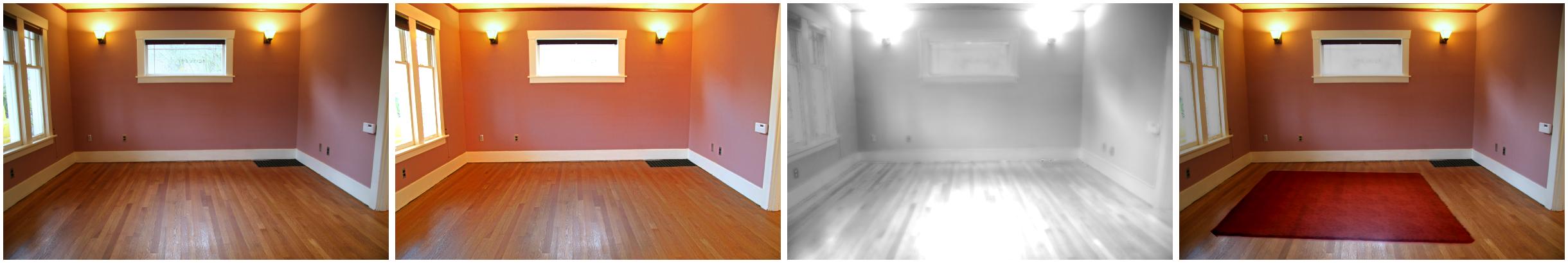} \\
    \includegraphics[width=1\linewidth,height=2.2cm]{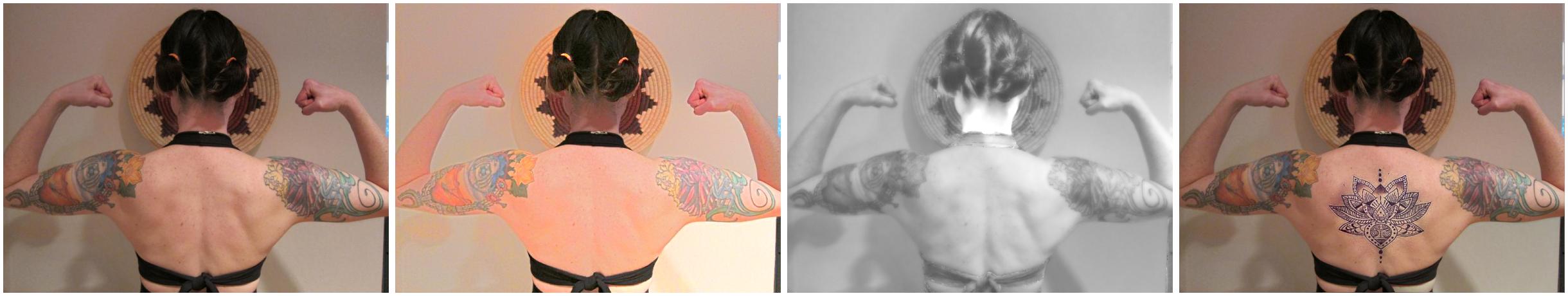} \\
    \includegraphics[width=1\linewidth,height=2.2cm]{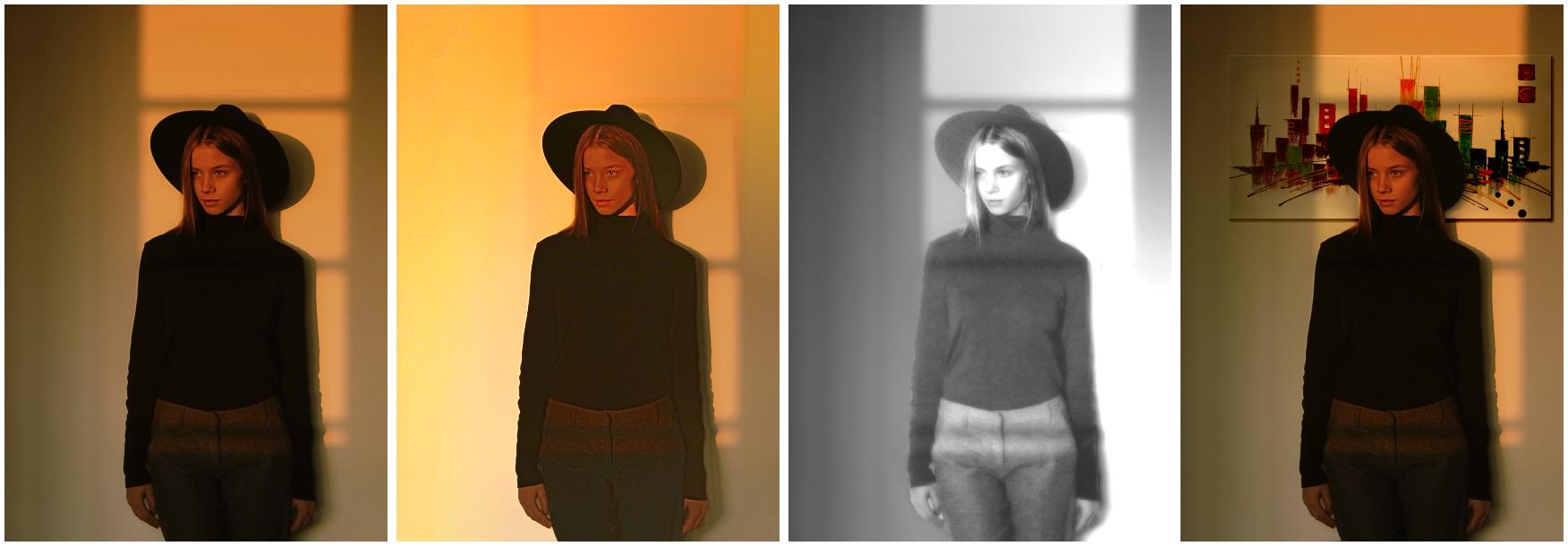} \\
    \includegraphics[width=1\linewidth,height=2.2cm]{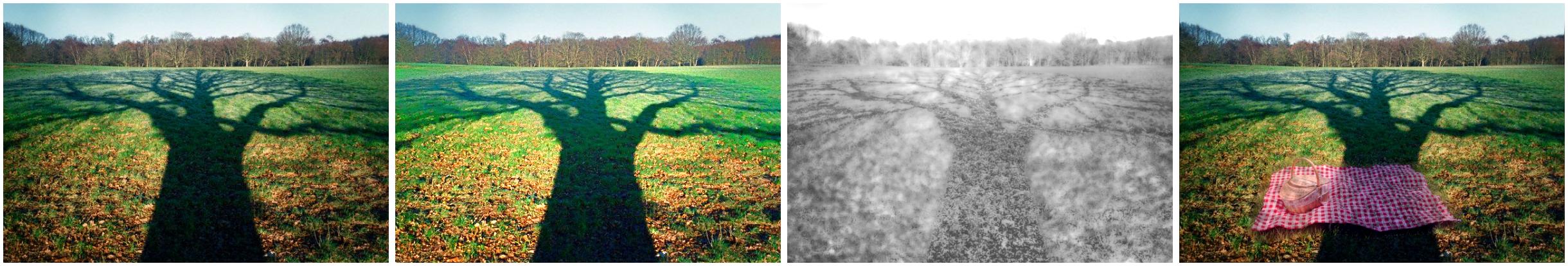} \\
  \end{minipage}
  \begin{minipage}{0.49\linewidth}
  \centering
    \includegraphics[width=1\linewidth,height=2.2cm]{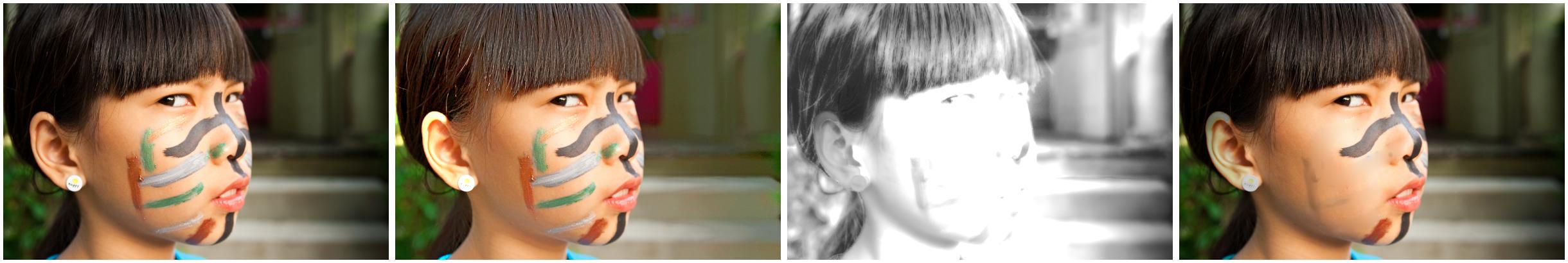} \\
    \includegraphics[width=1\linewidth,height=2.2cm]{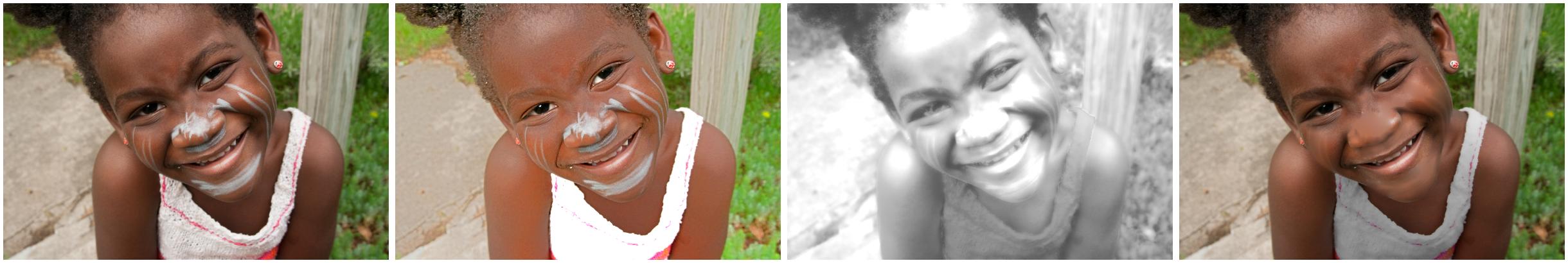} \\
    \includegraphics[width=1\linewidth,height=2.2cm]{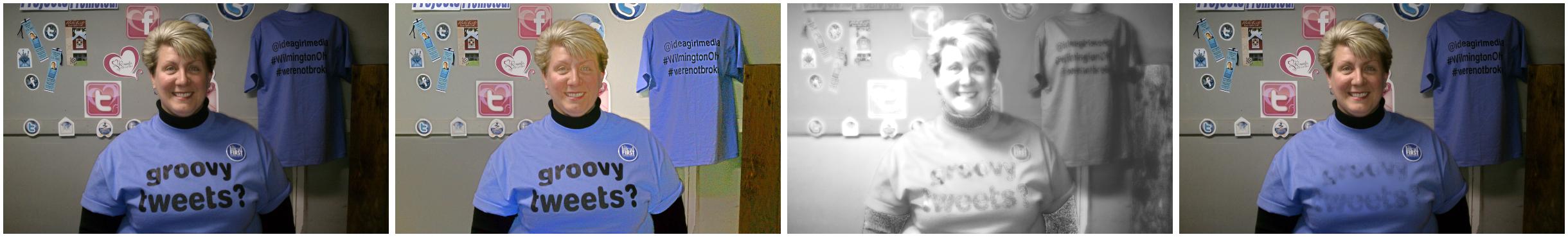} \\
    \includegraphics[width=1\linewidth,height=2.2cm]{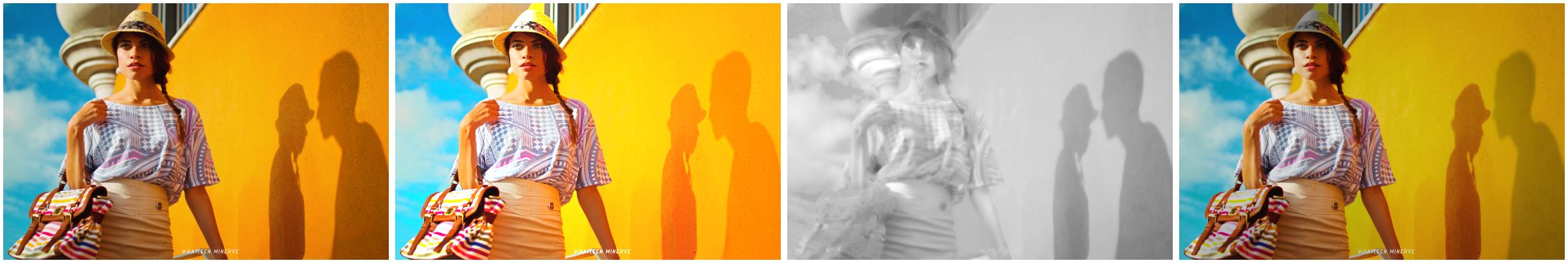} \\
    \rule{1\linewidth}{0.5pt}\vspace{2mm}
    \includegraphics[width=1\linewidth,height=2.2cm]{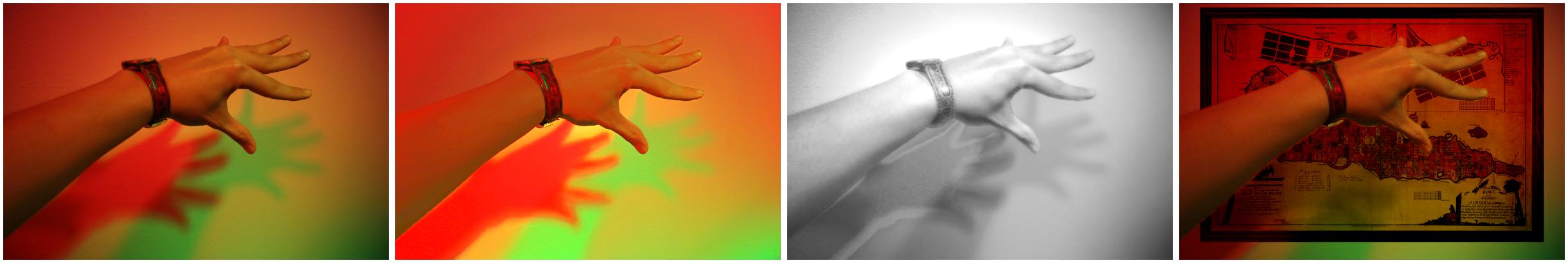} \\
    \includegraphics[width=1\linewidth,height=2.2cm]{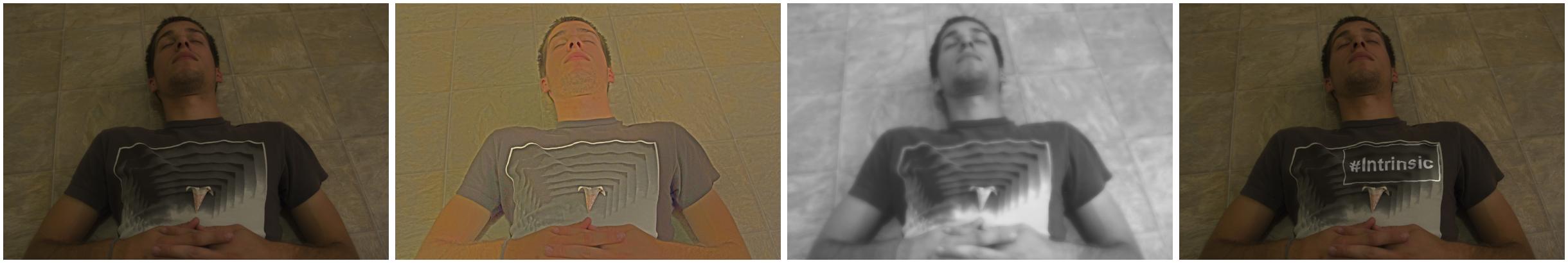} \\
    \includegraphics[width=1\linewidth,height=2.2cm]{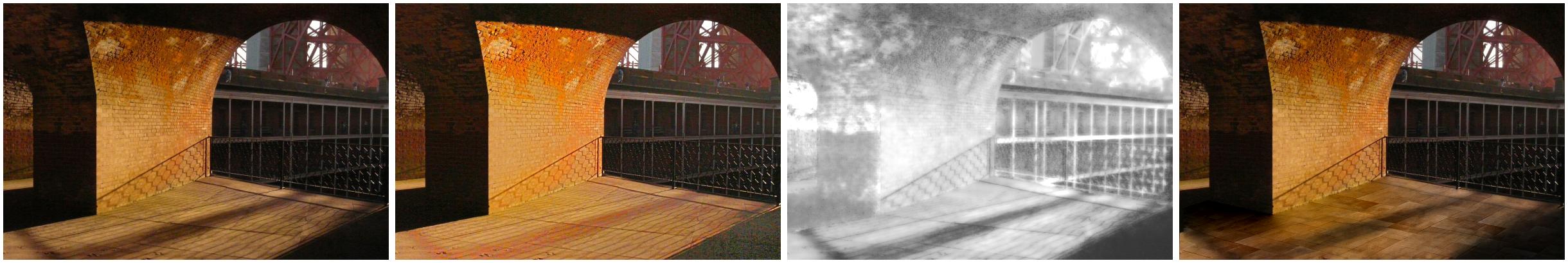} \\
    \includegraphics[width=1\linewidth,height=2.2cm]{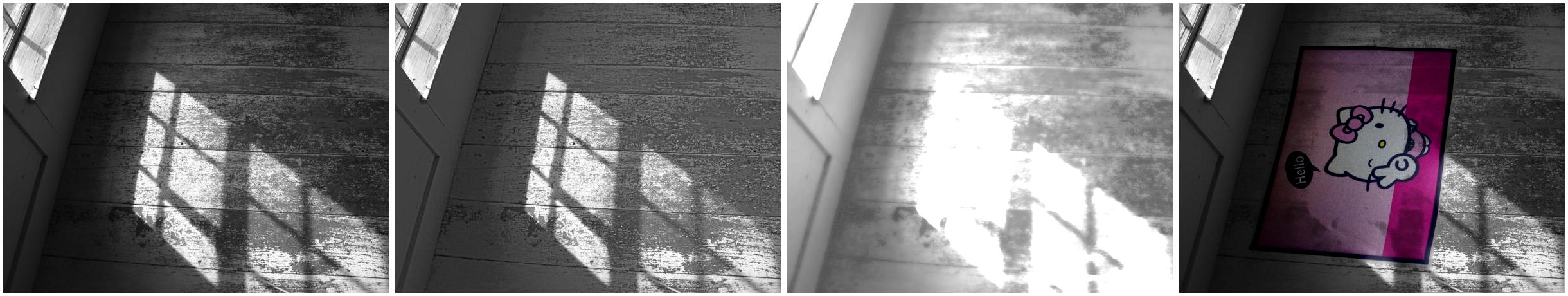} \\
  \end{minipage}
  \caption{\textbf{Image Editing Evaluation:} Our results on image editing applications from \cite{IIDstarreport}. Each row shows original image, reflectance, shading and the application result. The two sections from top respectively show results for texture removal and albedo replacement. The fourth row in each section shows the hard cases with sub-optimal decompositions due to sharp shadows or lighting.}
  \label{fig:starResults}
\end{figure*}

\subsection{Image Retexturing Applications}
Considering the limitation of quantitative evaluation metrics and in order to measure the effectiveness of the IID results for applications, \cite{IIDstarreport} proposed image editing based evaluation on set of few images over simple image manipulation applications like texture removal and albedo replacement. We show results of our method for these applications on the sample images in \fig \ref{fig:starResults}. 
We use the code, images and masks as provided by the authors and our merged components as inputs. As before, presented results are with fixed parameters and not tuned to the images or the application.
For comparison with other methods readers are requested to refer the supplementary material from the original authors\protect\footnote{\protect\url{https://perso.liris.cnrs.fr/nicolas.bonneel/intrinsicstar/supp_materials/image_editing/}}. As can been observed from the results, our method performs quite well in such real world application scenarios. 

\subsection{Image Relighting Applications}
Although our IID modeling is based on a simple light transport model (\sect \ref{sec:intro}), 
but here we show that we can approximate more complex lighting components for new relighting applications by combining the results from our two optimization stages.
The methods presented here are sample applications implemented as automatic but fixed parameter systems and require no user intervention during execution though an interactive interface an easily be incorporated.

\paragraph{\textbf{Editing Illumination Colour:}}
In order to change the illumination colour for image tone manipulation, we need to approximate illumination colour and light source regions.
As our original IID light transport model assumes monochromatic illumination, we approximate the low and high frequency components of illumination colour using the residue $C$.
We estimate the light source regions in the image by locating the pixels within the high percentile set in our merged shading component.
We change the intensity colour of shading component in CIELab space based upon the distance of pixels from the estimated light source regions.
This gives us a illumination colour modified shading component. 
Recombining this modified shading with reflectance gives us the tone manipulated image.

In \fig \ref{fig:appResults_2} we show illumination recolouring for several scenes using red and green tone modification.
Notice that the modified colour of the light source regions compared to the original image and the shading intensity based tone adjustment of the surrounding regions.
The objects farther away from the estimated light source regions, retain their original colour.
Unlike putting the entire image through a red or green filter, shading sensitive tone adjustment achieves a much more subtle and realistic tone manipulation effect. This illustrates a novel method for IID based illumination colour manipulation.

\paragraph{\textbf{Editing Illumination Intensity:}}
Given that the dynamic ranges of cameras and display devices are limited, some poorly or improperly lit images are excessively dark or bright in certain regions.
This is due to extreme intensity variation between bright light source regions \vs some dark unlit regions in a given scene.
Such images have Low Dynamic Range (LDR) of intensity compared to properly lit or intensity remapped images called High Dynamic Range (HDR) images.
We use our IID results to relight dark regions in a given image achieving the effect of single image LDR to HDR conversion.
Again we use the insight that our first stage results have smooth shading component whereas second stage has flat sparse reflectance component by design.
We can approximate directly and indirectly lit regions in the image as such regions are harder to decompose due to high intensity variations.
We extract additive and multiplicative residual information in the detailed components from their flat and smooth counterparts:
\[
E_1=mean((\rho - R),(S-\sigma)) \quad \text{and} \quad
\]
\[
E_2=mean(\frac{\rho}{R},\frac{S}{\sigma}).
\]
We add the Gaussian filtered estimates back to the original image and rescale the results between normal image intensity values for visualization.
By this we obtain the final well lit image as shown in \fig \ref{fig:appResults_1}.
Notice how the dark regions which were previously only indirectly lit are highlighted and intensity is maintained in the previously directly lit regions, achieving a shading sensitive intensity normalization. 
This presents a simple and novel method for IID based illumination intensity manipulation.

\begin{figure}[h]
\centering
 \includegraphics[width=0.96\linewidth,height=2.4cm]{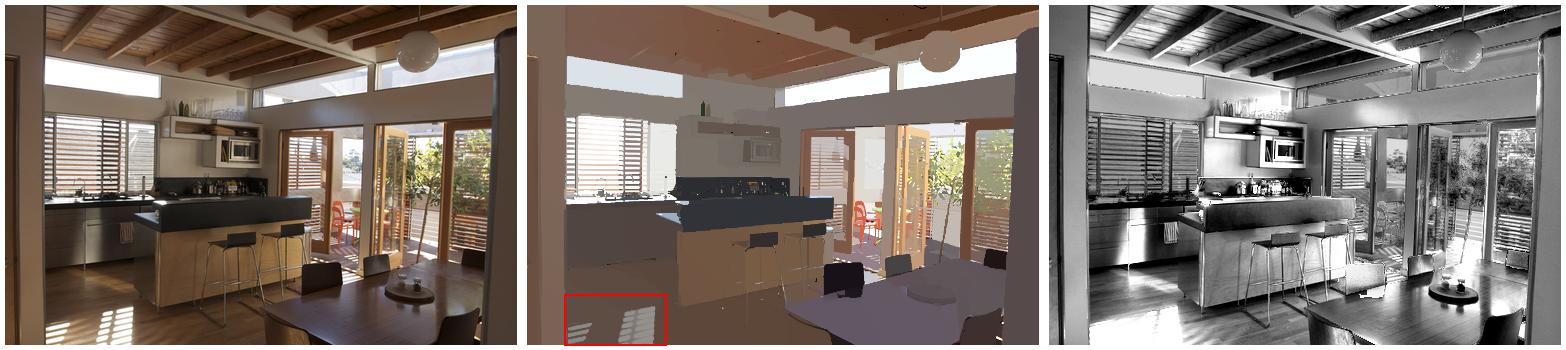} \\
 \includegraphics[width=0.96\linewidth,height=2.4cm]{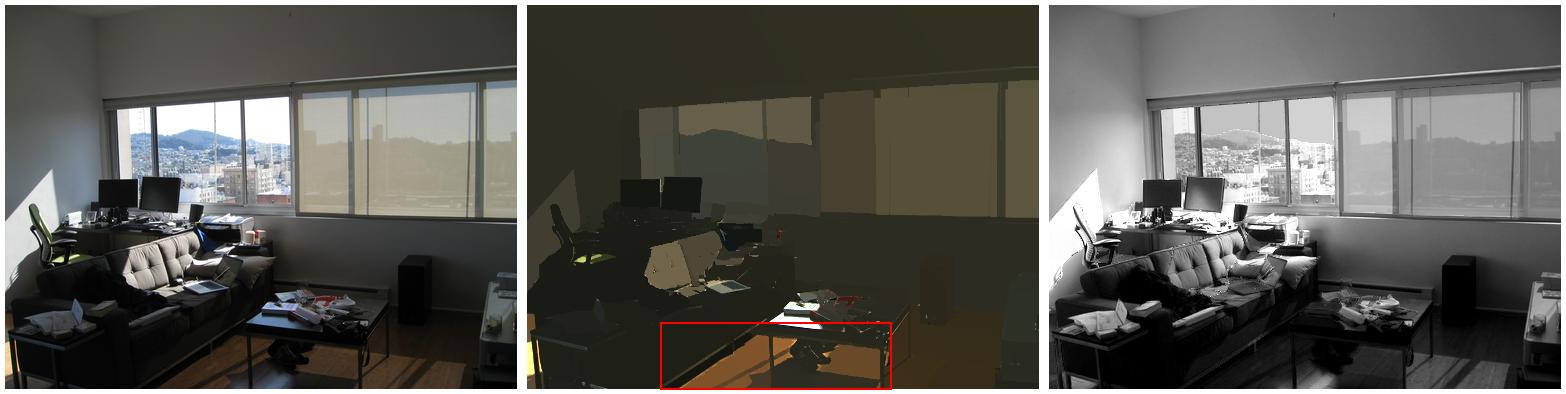} \\
 \includegraphics[width=0.96\linewidth,height=2.4cm]{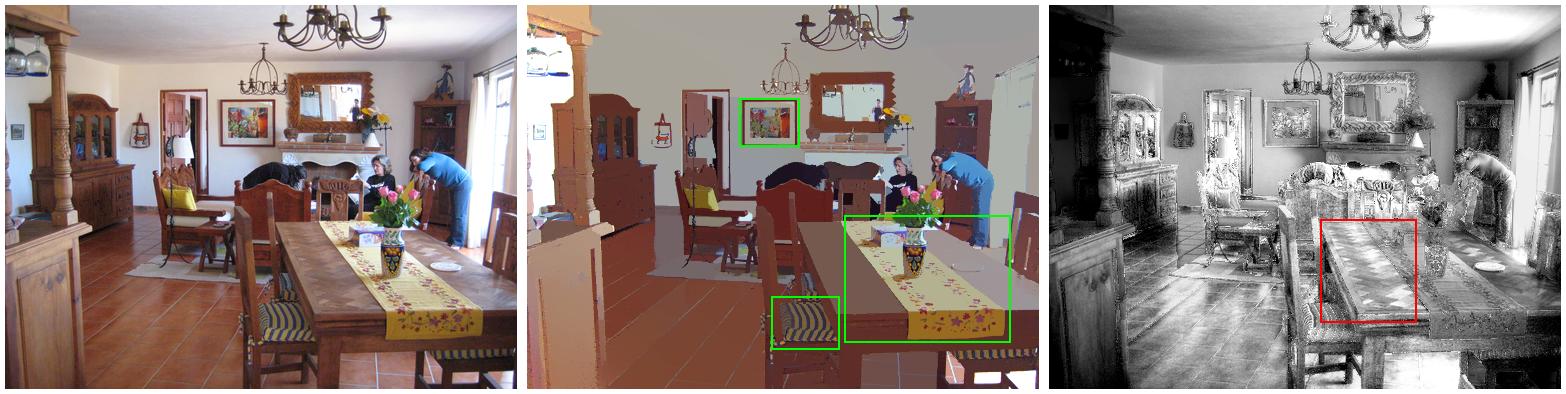} \\
\caption{\textbf{Failure cases:} Note incorrect decomposition in marked regions with challenging sharp highlights, shadows and fine textures in a colour similar to object.}
\label{fig:failCases}
\end{figure}
\section{Limitations and Future Work}
The \fig \ref{fig:failCases} shows a few failure scenarios of our proposed framework. An often observed challenging case is that of images with sharp shadow and highlight regions. 
Owing to the lack of depth data or some similar additional structural information, most single image IID methods struggle in this task of disambiguation of such gradients from sharp object boundaries. Yet another issue is distinguishing fine local textures in the same colour as object reflectance and lighting variation.
Our method is able to handle mid-level and large textures well due to our semantic priors but in a few cases such textures get decomposed into the shading layer.
Finer textures of colour similar to that of the object, persist in shading component due to ambiguity in differentiating local illumination changes with such textures (this is not an problem with differently coloured textures).
Notice how in the last image the textures on the table cloth are correctly decomposed into the reflectance layer but the textures on the wood owing to their similarity in colour are shifted to the shading component.
These issues are also observed in several other solutions \citep{IIDstarreport}.
Still our object semantic priors and alternating iterative model design leads to perceptually better decompositions for a large variety of scene and diverse lighting settings (\fig \ref{fig:internetResults}). 

Discounting the training time, deep learning based solutions generally run faster during testing in comparison to energy based optimization methods. 
Hence the unoptimized prototype implementation of our method is slower compared to other methods (few seconds \vs minutes) but this could be significantly improved with better implementation and parallelization.

In order to automatically assign the value of total number of iterations $k$ based on the lighting and scene complexity,
we would like to explore the problem of learning a performance metric for IID respecting both perceptual and quantitative assessment without ground truth information.
It would also be interesting to see the effect of explicitly introducing semantic information in current deep learning IID solutions.
We believe that properly encoding semantic and contextual information as an additional information, either as collated input, a separate network branch and/or as a loss function, would help improve the performance of the new IID deep learning solutions \citep{newBiIID,msrIID,snavelyIID_ECCV,snavelyIID}.
Additionally, it will also be interesting to see the utility of our and other recent IID solutions in novel applications like automatic video-editing, object insertion,
machine learning dataset augmentation, style-content disambiguation, \etc.
In future we would like to explore these questions in the context of IID and in the broader context of inverse rendering and inverse light transport research.

\section{Conclusion}
\label{sec:conclusions}
In this paper we present new priors which encode class agnostic weak object semantics using selective search and pre-trained region-based Convolutional Neural Network features.
We encode these priors by analyzing scene at three hierarchical context levels and use an integrated optimization framework for single image intrinsic image decomposition without requiring any additional optimization steps. 
Our system has two alternating optimization formulations with competing strategies: first focusing on shading smoothness and the second on reflectance sparsity.
We highlight the effectiveness of our strategy and semantic priors with supporting qualitative and quantitative experimentation and results.
We hope our work will draw attention of wider research community towards the utility of semantic priors and hierarchical analysis for the problem of intrinsic image decomposition and in the future will lead to better end-to-end deep learning architectures and optimization frameworks.
\begin{acknowledgements}
We would like to thank Tata Consultancy Services for supporting Saurabh Saini through Research Scholarship Program (TCS RSP) during the project.
\end{acknowledgements}

\bibliographystyle{spbasic}      
\bibliography{spiid}             

\begin{thebibliography}{60}
\providecommand{\natexlab}[1]{#1}
\providecommand{\url}[1]{{#1}}
\providecommand{\urlprefix}{URL }
\expandafter\ifx\csname urlstyle\endcsname\relax
  \providecommand{\doi}[1]{DOI~\discretionary{}{}{}#1}\else
  \providecommand{\doi}{DOI~\discretionary{}{}{}\begingroup
  \urlstyle{rm}\Url}\fi
\providecommand{\eprint}[2][]{\url{#2}}

\bibitem[{Arbel\'{a}ez et~al.(2014)Arbel\'{a}ez, Pont-Tuset, Barron, Marques,
  and Malik}]{MCG}
Arbel\'{a}ez P, Pont-Tuset J, Barron J, Marques F, Malik J (2014) Multiscale
  combinatorial grouping. CVPR

\bibitem[{Barron(2012)}]{barronIID_CVPR12}
Barron JT (2012) Shape, albedo, and illumination from a single image of an
  unknown object. CVPR

\bibitem[{Barron and Malik(2012)}]{barronIID_eccv12}
Barron JT, Malik J (2012) Color constancy, intrinsic images, and shape
  estimation. ECCV

\bibitem[{Barron and Malik(2013)}]{barronIID_RGBD}
Barron JT, Malik J (2013) Intrinsic scene properties from a single rgb-d image.
  CVPR

\bibitem[{Barron and Malik(2015)}]{barronIID_PAMI}
Barron JT, Malik J (2015) Intrinsic scene properties from a single rgb-d image.
  TPAMI

\bibitem[{Bell et~al.(2014)Bell, Bala, and Snavely}]{balaIID}
Bell S, Bala K, Snavely N (2014) Intrinsic images in the wild. ACM Transactions
  on Graphics (SIGGRAPH) 33(4)

\bibitem[{Bi et~al.(2015)Bi, Han, and Yu}]{biIID}
Bi S, Han X, Yu Y (2015) An \emph{L}\({}_{\mbox{1}}\) image transform for
  edge-preserving smoothing and scene-level intrinsic decomposition. ACM
  Transactions on Graphics 34(4)

\bibitem[{Bi et~al.(2018)Bi, Kalantari, and Ramamoorthi}]{newBiIID}
Bi S, Kalantari NK, Ramamoorthi R (2018) {Deep Hybrid Real and Synthetic
  Training for Intrinsic Decomposition}. EGSR

\bibitem[{Bonneel et~al.(2014)Bonneel, Sunkavalli, Tompkin, Sun, Paris, and
  Pfister}]{bonneelIID}
Bonneel N, Sunkavalli K, Tompkin J, Sun D, Paris S, Pfister H (2014)
  Interactive intrinsic video editing. ACM Transactions on Graphics (SIGGRAPH
  Asia) 33(6)

\bibitem[{Bonneel et~al.(2017)Bonneel, Kovacs, Paris, and Bala}]{IIDstarreport}
Bonneel N, Kovacs B, Paris S, Bala K (2017) Intrinsic decompositions for image
  editing. Computer Graphics Forum (Eurographics State of the Art Reports)
  36(2)

\bibitem[{Bousseau et~al.(2009)Bousseau, Paris, and Durand}]{bousseauIID}
Bousseau A, Paris S, Durand F (2009) User assisted intrinsic images. ACM
  Transactions on Graphics (SIGGRAPH Asia) 28(5)

\bibitem[{Butler et~al.(2012)Butler, Wulff, Stanley, and Black}]{mpiSintel}
Butler DJ, Wulff J, Stanley GB, Black MJ (2012) A naturalistic open source
  movie for optical flow evaluation. ECCV

\bibitem[{Carroll et~al.(2011)Carroll, Ramamoorthi, and
  Agrawala}]{ramamoorthiRelight}
Carroll R, Ramamoorthi R, Agrawala M (2011) Illumination decomposition for
  material recoloring with consistent interreflections. ACM Transactions on
  Graphics 30(4)

\bibitem[{Chang et~al.(2014)Chang, Cabezas, and Fisher}]{bayesianNonParamIID}
Chang J, Cabezas R, Fisher JW (2014) Bayesian nonparametric intrinsic image
  decomposition. ECCV

\bibitem[{Chen and Koltun(2013)}]{koltunIID}
Chen Q, Koltun V (2013) A simple model for intrinsic image decomposition with
  depth cues. ICCV

\bibitem[{Deng et~al.(2009)Deng, Dong, Socher, jia Li, Li, and
  Fei-fei}]{Imagenet}
Deng J, Dong W, Socher R, jia Li L, Li K, Fei-fei L (2009) Imagenet: A
  large-scale hierarchical image database. CVPR

\bibitem[{Donahue et~al.(2014)Donahue, Jia, Vinyals, Hoffman, Zhang, Tzeng, and
  Darrell}]{donahueDecaf}
Donahue J, Jia Y, Vinyals O, Hoffman J, Zhang N, Tzeng E, Darrell T (2014)
  Decaf: A deep convolutional activation feature for generic visual
  recognition. ICML

\bibitem[{Duch\^{e}ne et~al.(2015)Duch\^{e}ne, Riant, Chaurasia, Moreno,
  Laffont, Popov, Bousseau, and Drettakis}]{multiviewOutdoorRelight}
Duch\^{e}ne S, Riant C, Chaurasia G, Moreno JL, Laffont PY, Popov S, Bousseau
  A, Drettakis G (2015) Multiview intrinsic images of outdoors scenes with an
  application to relighting. ACM Transactions on Graphics 34(5)

\bibitem[{Fan et~al.(2018)Fan, Yang, Hua, Chen, and Wipf}]{msrIID}
Fan Q, Yang J, Hua G, Chen B, Wipf D (2018) Revisiting deep intrinsic image
  decompositions. CVPR

\bibitem[{Garces et~al.(2012)Garces, Munoz, Lopez-Moreno, and
  Gutierrez}]{clusteringIID}
Garces E, Munoz A, Lopez-Moreno J, Gutierrez D (2012) Intrinsic images by
  clustering. Computer Graphics Forum (Proc EGSR) 31(4)

\bibitem[{Gehler et~al.(2011)Gehler, Rother, Kiefel, Zhang, and
  Sch\"{o}lkopf}]{gehlerIID}
Gehler PV, Rother C, Kiefel M, Zhang L, Sch\"{o}lkopf B (2011) Recovering
  intrinsic images with a global sparsity prior on reflectance. NeurIPS

\bibitem[{Girshick et~al.(2014)Girshick, Donahue, Darrell, and Malik}]{rcnn}
Girshick R, Donahue J, Darrell T, Malik J (2014) Rich feature hierarchies for
  accurate object detection and semantic segmentation. CVPR

\bibitem[{Goldstein and Osher(2009)}]{splitBregman}
Goldstein T, Osher S (2009) The split bregman method for l1-regularized
  problems. SIAM J Img Sci 2(2)

\bibitem[{Grosse et~al.(2009)Grosse, Johnson, Adelson, and
  Freeman}]{mitDataset}
Grosse R, Johnson MK, Adelson EH, Freeman WT (2009) Ground-truth dataset and
  baseline evaluations for intrinsic image algorithms. ICCV

\bibitem[{Hosang et~al.(2014)Hosang, Benenson, and
  Schiele}]{detectionProposalsSurvey}
Hosang J, Benenson R, Schiele B (2014) How good are detection proposals,
  really? BMVC

\bibitem[{Jeon et~al.(2014)Jeon, Cho, Tong, and Lee}]{jeonIID}
Jeon J, Cho S, Tong X, Lee S (2014) Intrinsic image decomposition using
  structure-texture separation and surface normals. ECCV

\bibitem[{Karsch et~al.(2014)Karsch, Liu, and Kang}]{karschDepth}
Karsch K, Liu C, Kang SB (2014) Depth transfer: Depth extraction from video
  using non-parametric sampling. TPAMI 36(11)

\bibitem[{Khosla et~al.(2012)Khosla, Zhou, Malisiewicz, Efros, and
  Torralba}]{khoslaDatasetBias}
Khosla A, Zhou T, Malisiewicz T, Efros A, Torralba A (2012) Undoing the damage
  of dataset bias. ECCV

\bibitem[{Kim et~al.(2016)Kim, Park, Sohn, and Lin}]{stephenlinIID}
Kim S, Park K, Sohn K, Lin S (2016) Unified depth prediction and intrinsic
  image decomposition from a single image via joint convolutional neural
  fields. ECCV

\bibitem[{Kong et~al.(2014)Kong, Gehler, and Black}]{videoIID}
Kong N, Gehler PV, Black MJ (2014) Intrinsic video. ECCV

\bibitem[{Kovacs et~al.(2017)Kovacs, Bell, Snavely, and Bala}]{sawIID}
Kovacs B, Bell S, Snavely N, Bala K (2017) Shading annotations in the wild.
  CVPR

\bibitem[{Kwatra et~al.(2012)Kwatra, Han, and Dai}]{kwatraShadowRemoval}
Kwatra V, Han M, Dai S (2012) Shadow removal for aerial imagery by information
  theoretic intrinsic image analysis. International Conference on Computational
  Photography (ICCP)

\bibitem[{Laffont et~al.(2012)Laffont, Bousseau, Paris, Durand, and
  Drettakis}]{photocollectionIID}
Laffont PY, Bousseau A, Paris S, Durand F, Drettakis G (2012) Coherent
  intrinsic images from photo collections. ACM Transactions on Graphics 31(6)

\bibitem[{Laffont et~al.(2013)Laffont, Bousseau, and Drettakis}]{multiviewIID}
Laffont PY, Bousseau A, Drettakis G (2013) Rich intrinsic image decomposition
  of outdoor scenes from multiple views. IEEE Transactions on Visualization and
  Computer Graphics 19(2)

\bibitem[{Land and McCann(1971)}]{Land71}
Land EH, McCann JJ (1971) Lightness and retinex theory. J Opt Soc Am 61(1)

\bibitem[{Levin et~al.(2006)Levin, Lischinski, and Weiss}]{laplacianMatting}
Levin A, Lischinski D, Weiss Y (2006) A closed form solution to natural image
  matting. CVPR

\bibitem[{Li and Brown(2014)}]{layerSeparation}
Li Y, Brown MS (2014) Single image layer separation using relative smoothness.
  CVPR

\bibitem[{Li and Snavely(2018{\natexlab{a}})}]{snavelyIID_ECCV}
Li Z, Snavely N (2018{\natexlab{a}}) Cgintrinsics: Better intrinsic image
  decomposition through physically-based rendering. ECCV

\bibitem[{Li and Snavely(2018{\natexlab{b}})}]{snavelyIID}
Li Z, Snavely N (2018{\natexlab{b}}) Learning intrinsic image decomposition
  from watching the world. CVPR

\bibitem[{Liu et~al.(2011)Liu, Yuen, and Torralba}]{siftFlow}
Liu C, Yuen J, Torralba A (2011) Sift flow: Dense correspondence across scenes
  and its applications. TPAMI 33(5)

\bibitem[{Liu et~al.(2008)Liu, Wan, Qu, Wong, Lin, Leung, and
  Heng}]{iidColorization}
Liu X, Wan L, Qu Y, Wong TT, Lin S, Leung CS, Heng PA (2008) Intrinsic
  colorization. ACM Transactions on Graphics 27(5)

\bibitem[{Long et~al.(2014)Long, Zhang, and Darrell}]{convNetsCorrespondence}
Long J, Zhang N, Darrell T (2014) Do convnets learn correspondence? NeurIPS

\bibitem[{Marschner(1998)}]{MarschnerThesis}
Marschner SR (1998) Inverse rendering for computer graphics. PhD thesis

\bibitem[{Narihira et~al.(2015{\natexlab{a}})Narihira, Maire, and
  Yu}]{narihiraIID}
Narihira T, Maire M, Yu SX (2015{\natexlab{a}}) Direct intrinsics: Learning
  albedo-shading decomposition by convolutional regression. ICCV

\bibitem[{Narihira et~al.(2015{\natexlab{b}})Narihira, Maire, and
  Yu}]{NarihiraSplit}
Narihira T, Maire M, Yu SX (2015{\natexlab{b}}) Learning lightness from human
  judgement on relative reflectance. CVPR

\bibitem[{Nestmeyer and Gehler(2017)}]{filteringIID}
Nestmeyer T, Gehler PV (2017) Reflectance adaptive filtering improves intrinsic
  image estimation. CVPR

\bibitem[{Ramamoorthi and Hanrahan(2001)}]{RamamoorthiInverseRendering}
Ramamoorthi R, Hanrahan P (2001) A signal-processing framework for inverse
  rendering. Proceedings of the 28th Annual Conference on Computer Graphics and
  Interactive Techniques

\bibitem[{Saini and Narayanan(2018)}]{SPIID}
Saini S, Narayanan PJ (2018) Semantic priors for intrinsic image decomposition.
  BMVC

\bibitem[{Saini et~al.(2016)Saini, Sakurikar, and Narayanan}]{fstackIID}
Saini S, Sakurikar P, Narayanan PJ (2016) Intrinsic image decomposition using
  focal stacks. Proceedings of the Tenth Indian Conference on Computer Vision,
  Graphics and Image Processing

\bibitem[{Sharif~Razavian et~al.(2014)Sharif~Razavian, Azizpour, Sullivan, and
  Carlsson}]{astoundingCNN}
Sharif~Razavian A, Azizpour H, Sullivan J, Carlsson S (2014) Cnn features
  off-the-shelf: An astounding baseline for recognition. CVPR Workshops

\bibitem[{Shelhamer et~al.(2015)Shelhamer, Barron, and Darrell}]{shelhamerIID}
Shelhamer E, Barron JT, Darrell T (2015) Scene intrinsics and depth from a
  single image. ICCV Workshops

\bibitem[{Shen et~al.(2013)Shen, Yeo, and Hua}]{shenIID}
Shen L, Yeo C, Hua BS (2013) Intrinsic image decomposition using a sparse
  representation of reflectance. TPAMI 35(12)

\bibitem[{Tappen et~al.(2005)Tappen, Freeman, and Adelson}]{tappenIID}
Tappen MF, Freeman WT, Adelson EH (2005) Recovering intrinsic images from a
  single image. TPAMI 27(9)

\bibitem[{Torralba and Efros(2011)}]{torralbaDatasetBias}
Torralba A, Efros AA (2011) Unbiased look at dataset bias. CVPR

\bibitem[{Vineet et~al.(2013)Vineet, Rother, and Torr}]{vibhavIID}
Vineet V, Rother C, Torr PHS (2013) Higher order priors for joint intrinsic
  image, objects, and attributes estimation. In: NeurIPS

\bibitem[{Weiss(2001)}]{weissIID}
Weiss Y (2001) Deriving intrinsic images from image sequences. ICCV

\bibitem[{Yosinski et~al.(2014)Yosinski, Clune, Bengio, and
  Lipson}]{bengioTransferableFeatures}
Yosinski J, Clune J, Bengio Y, Lipson H (2014) How transferable are features in
  deep neural networks? NeurIPS

\bibitem[{Zhao et~al.(2012)Zhao, Tan, Dai, Shen, Wu, and Lin}]{zhaoIID}
Zhao Q, Tan P, Dai Q, Shen L, Wu E, Lin S (2012) A closed-form solution to
  retinex with nonlocal texture constraints. TPAMI 34(7)

\bibitem[{Zhou et~al.(2015)Zhou, Krahenbuhl, and Efros}]{efrosIID}
Zhou T, Krahenbuhl P, Efros AA (2015) Learning data-driven reflectance priors
  for intrinsic image decomposition. ICCV

\bibitem[{Zoran et~al.(2015)Zoran, Isola, Krishnan, and Freeman}]{freemanIID}
Zoran D, Isola P, Krishnan D, Freeman WT (2015) Learning ordinal relationships
  for mid-level vision. ICCV

\end{thebibliography}

\end{document}